\documentclass[12pt,letterpaper]{article}

\newcommand{\blind}{1}

\usepackage{latexsym,multirow}
\usepackage{amssymb,amsmath, bm}
\usepackage{graphicx}
 
\usepackage{lmodern}

\usepackage{natbib}
\usepackage{bibunits}
\usepackage{color}
\usepackage[pdftex, bookmarksopen=true, bookmarksnumbered=true,
pdfstartview=FitH, breaklinks=true, urlbordercolor={0 1 0}, citebordercolor={0 0 1}]{hyperref}

\DeclareMathAlphabet{\mathpzc}{OT1}{pzc}{m}{it}
\usepackage{subfigure}
\usepackage{placeins}
\usepackage{comment}

\usepackage{pifont}

\usepackage{dcolumn}
\newcolumntype{.}{D{.}{.}{-1}}
\newcolumntype{d}[1]{D{.}{.}{#1}}

\usepackage{tikz}
\usepackage[latin1]{inputenc}
\usetikzlibrary{shapes,arrows,trees,fit,positioning}

\usepackage{theorem}
\theoremstyle{plain}
\theoremheaderfont{\scshape}

\renewcommand{\Pr}{\textrm{Pr}}

\DeclareMathOperator{\sgn}{sgn}

\usepackage{rotating}

\usepackage{arydshln}

\usepackage[compact]{titlesec}
\usepackage[toc,page,header]{appendix}
\usepackage{minitoc}

\usepackage[ruled,vlined]{algorithm2e}

\begin{document}
\fontsize{12pt}{20 pt plus 1 pt minus 0.5 pt}\selectfont

%Special instructions to get rid of the annoying vector notation
\newcommand\vm[1]{% Vector or matrix
\bm{\mathrm{#1}}}
% === new commands ===
\newcommand\ud{\mathrm{d}}
\newcommand\dist{\buildrel\rm d\over\sim}
\newcommand\ind{\stackrel{\rm indep.}{\sim}}
\newcommand\iid{\stackrel{\rm i.i.d.}{\sim}}
\newcommand\logit{{\rm logit}}
\newcommand\convd{\stackrel{\rm d}{\rightarrow}}
\newcommand\convp{\stackrel{\rm p}{\rightarrow}}
\newcommand\convu{\stackrel{\rm u}{\rightarrow}}
\renewcommand\r{\right}
\renewcommand\l{\left}
\newcommand\pre{{(t-1)}}
\newcommand\cur{{(t)}}
\newcommand\cA{\mathcal{A}}
\newcommand\bone{\mathbf{1}}
\newcommand\E{\mathbb{E}}
\newcommand\Var{{\rm Var}}
\newcommand\Cov{{\rm Cov}}
\newcommand\Corr{{\rm Corr}}
\newcommand\med{{\rm med}}
\newcommand\cC{\mathcal{C}}
\newcommand\cD{\mathcal{D}}
\newcommand\cJ{\mathcal{J}}
\newcommand\cN{\mathcal{N}}
\newcommand\cP{\mathcal{P}}
\newcommand\cT{\mathcal{T}}
\newcommand\cX{\mathcal{X}}
\newcommand\wX{\widetilde{X}}
\newcommand\wT{\widetilde{T}}
\newcommand\wY{\widetilde{Y}}
\newcommand\wZ{\widetilde{Z}}
\newcommand\oX{\overline{X}}
\newcommand\oT{\overline{T}}
\newcommand\oY{\overline{Y}}
\newcommand\bX{\mathbf{X}}
\newcommand\bB{\mathbf{B}}
\newcommand\bD{\mathbf{D}}
\newcommand\bM{\mathbf{M}}
\newcommand\bH{\mathbf{H}}
\newcommand\bI{\mathbf{I}}
\newcommand\bG{\mathbf{G}}
\newcommand\bR{\mathbf{R}}
\newcommand\bS{\mathbf{S}}
\newcommand\bV{\mathbf{V}}
\newcommand\bW{\mathbf{W}}
\newcommand\plim{\textrm{plim}\;}
\newcommand\supp{\textrm{supp}\;}
\newcommand\card{\textrm{card}\;}
\newcommand\expit{\textrm{expit}\;}
\newcommand\sign{\textrm{sign}\;}

\newcommand{\argmax}{\operatornamewithlimits{argmax}}
\newcommand{\argmin}{\operatornamewithlimits{argmin}}

\newcommand\spacingset[1]{\renewcommand{\baselinestretch}%
{#1}\small\normalsize}

\spacingset{1.25}

\newcommand{\tit}{Estimation and Inference on Nonlinear and Heterogeneous Effects}

%%%%%%%%%%%%%%%%%%%%%%%%%%%%%%%%%%%%%%%%%%%%%%%%%%%%%%%%%%%%%%%%%%%%%%%%

\if1\blind

{\title{\bf \tit \thanks{We thank Max Gopelrud, Kosuke Imai, Lucas Janson, Shiro Kuriwaki, Lihua Lei, Lisa McKay, Max Farrell, and Brandon Stewart for comments on this paper. A previous unpublished paper, ``The Method of Direct Estimation'' (2017), worked towards the approach to point estimates used in this paper. On this prior work, we would like to thank Peter Aronow, Scott de Marchi, James Fowler, Andrew Gelman, Max Gopelrud, Kosuke Imai, Gary King, Shiro Kuriwaki, John Londregan, Chris Lucas, Walter Mebane,  Rich Nielsen, Molly Roberts, Brandon Stewart, Aaron Strauss, Rocio Titiunik, Tyler VanderWeele, Teppei Yamamoto, Soichiro Yamauchi, and Xiang Zhou, as well as the participants at the Quantitative Social Science Seminar at Princeton,  Yale Research Design and Causal Inference seminar, Empirical Implications of Theoretical Models 2018 workshop, and Harvard Applied Statistics workshop.}}

\author{Marc Ratkovic\thanks{Assistant Professor, Department of
  Politics, Princeton University, Princeton NJ 08544. Phone:
  608-658-9665, Email:
  \href{mailto:ratkovic@princeton.edu}{ratkovic@princeton.edu},
  URL:
  \href{http://scholar.princeton.edu/ratkovic}{http://scholar.princeton.edu/ratkovic}}\hspace{.1in}
Dustin Tingley\thanks{Professor of Government, Harvard University, Email:
  \href{mailto:dtingley@gov.harvard.edu}{dtingley@gov.harvard.edu},
  URL:
  \href{http://scholar.harvard.edu/dtingley}{scholar.harvard.edu/dtingley} } }

\date{\today}

\maketitle

}\fi

\pdfbookmark[1]{Title Page}{Title Page}

\thispagestyle{empty}

\begin{abstract}

Multiple regression has been the go-to method for data analysis for generations of scholars due to its transparency, interpretability, and desirable theoretical properties. However, the method's simplicity precludes the discovery of complex heterogeneities in the data. We introduce the Method of Direct Estimation and Inference (MDEI) that embraces these potential complexities, is interpretable, has desirable theoretical guarantees, and, unlike some existing methods, returns appropriate uncertainty estimates. The proposed method uses a machine learning regression methodology to estimate the observation-level effect of a treatment variable. Importantly, we introduce a robust approach to uncertainty estimates. We provide simulation evidence and an application illustrating the performance of the method.

\end{abstract}

\spacingset{1.9}

\clearpage
\pagenumbering{arabic} 
\section{Introduction}

Data analysis in much of political science and other social sciences is often synonymous with multiple linear regression.  In this project, we assume the researcher confronts an outcome variable,  a  treatment variable of central interest, and a set of background ``control''/``confounding'' variables that characterizes each observation's covariate profile. The usefulness of multiple regression in this context depends in part on correctly modelling the influence of the treatment variable while adjusting for confounding effects of other variables. Typical regression strategies commonly ignore complexity in the data, such as the heterogeneous effect of the treatment across the sample (a treatment by covariate interaction), or they assume all effects are linear (both the treatment and confounders). Departures from typical practice tend to be ad hoc, with maybe one interaction or non-linearity considered. While methods have been introduced for moving beyond multiple regression for finding nonlinearities and interactions, estimating these nonlinearities and interactions is not the same as also returning appropriate uncertainty estimates.  

We introduce a novel method for finding nonlinear and heterogeneous effects, and focus on how to appropriate calculate uncertainty in these settings. We propose the Method of Direct Estimation and Inference (MDEI)--that embraces these heterogeneities and nonlinearities while still returning appropriate uncertainty estimates on effects. We focus largely on the case of a continuous treatment variable but also consider the binary case. As with much work in the causal inference literature \citep{aronowfoundations,imbens2015causal,ho:imai:king:stua:07,holl:86} we focus on reducing the role of modelling assumptions. However, our approach minimizes the role of assumptions in estimating both point estimates and uncertainty estimates.

%As a result researchers implicitly make many strong assumptions about the data. 

%This is in contrast with advances in the causal inference literature,  which works to reduce the impact of modeling choices on causal estimates . Furthermore, ignoring potentially interesting interactions and nonlinearities can miss adding additional nuance to a researcher's insights \citep[e.g.][]{Hill2014,  beck1998beyond, Beck2000}.   

Estimating the treatment effect requires developing a method that can return an accurate point estimate of the estimand. This requires flexibly modeling two things: 1) the background covariates/confounders and 2) the treatment effect. Using a novel machine learning method we flexibly adjust for background covariates while also allowing for substantial flexibility in the effect of the treatment on the outcome.
%Marc, I removed this footnote from the intro. We cover this in the body sufficiently, right?%\footnote{Recent work has advocated using machine learning methods to remove the effect of confounders from the outcome and treatment, then running a regression on the residualized outcome of the residualized treatment \citep[e.g.,][]{chern:chet:2018}.  While this can help reduce the model's dependence on how covariates enter the model, it does not capture underlying heterogeneities (e.g., interactions between the treatment and covariates) or nonlinear effects of the treatment that may be of interest to the researcher.}  

The next step--and crucial to this paper--is to generate uncertainty estimates and confidence bands for the results. This is straightforward when there are strong parametric modelling assumptions in place, like with multiple linear regression. The task is much more challenging when we want to allow for more complicated relationships that we do not specify ex-ante. Our goal is to generate a confidence band around any uncovered nonlinearity that will allow us to assess how the estimated curve relates to the true curve. Thinking about uncertainty in this setting requires differentiating between inference at a particular point and inference over a curve.  For this, we estimate a  confidence band, meaning we expect the confidence band for our marginal effect to cover the true curve at some proportion, say 90\%, of the data.  Doing this allows us to use the uncertainty measure around the curve to deduce features of the true underlying curve.

In summary, the MDEI framework provides flexible and reliable estimation \textit{and} inference.  The method consists of two parts. First, we estimate the instantaneous causal effect, which is the observation-level marginal effect of the treatment on the outcome \citep[e.g.][]{Stolzenberg80}.  Just as a regression coefficient is interpreted as an average marginal effect over the sample, the instantaneous causal effect is interpreted as the ``slope'' at a given observation, given the values of observed pre-treatment variables. To perform this estimate, MDEI advances recent machine learning methods by implementing a flexible, nonparametric regression  to model the treatment effect.  The model can detect a wide class of nonlinear and treatment/covariate interactions.  Here, we build on an approach known in the econometrics literature known as seminonparametrics \citep[see e.g.,][]{chen:07} where we model the contributions of both  the covariates treatment/covariate interactions using flexible machine learning methods.  

Second, we introduce a confidence band on uncovered nonlinearities and heterogeneities that the researcher can use to assess whether a given effect reflects a systematic pattern in the data.  The curve has the {\it average coverage property} \citep{Nychka1988,wass:06} that the band will contain the true treatment effect at some chosen proportion, say 90\% or 95\%, of the observed data.  In constructing such a curve, we rely on {\it conformal inference} \citep{lei:was:14}. As discussed below, conformal inference provides a data driven, rather than assumption driven, approach to calculating uncertainty estimates on predicted values. We extend the method from predicted values to estimating the marginal effect of the treatment on the outcome at each point.

The MDEI framework draws on tools and ideas that might be new for many readers in political science. Throughout the paper we try to introduce these ideas in an accessible manner and refer readers to a more technical appendix. While the MDEI framework introduced here is new, we relate our approach to existing methods where relevant.  

The paper proceeds as follows. Section~\ref{sec:challenge} lays out the challenge of estimating and conducting inference on treatment effects without relying on simplifying assumptions about how the treatment impacts the outcome variable. In Section~\ref{sec:classes} we build up from a model where these assumptions are very strong to the our model where these assumptions are very minimal. Section~\ref{sec:uncertainty}  introduces our approach to inference in this setting where we develop a confidence band utilizing a  conformal inference approach. Section~\ref{sec:estimation} then introduces MDEI and shows how we estimate both point estimates and uncertainty estimates.  Section~\ref{sec:illustrative} provides simulations to illustrate our approach while the Appendix compares the performance of MDEI to other cutting edge approaches. Section~\ref{sec:example} shows MDEI in action with an applied example. Section~\ref{sec:conclusion} concludes. Throughout we discuss related research. We include an extensive online technical appendix. 

\section{The Estimation and Uncertainty Challenge}
\label{sec:challenge}

Consider the familiar regression model,
\begin{align*}
Y = \theta T + X^\top \gamma + \epsilon;\;\; \E(\epsilon|T,X)=0
\end{align*}
with outcome $Y$, a variable of theoretical interest $T$, a vector of additional background variables $X$, and an error term $\epsilon$ that is assumed to be mean-independent of the treatment and background variables.   This model is adopted for several reasons.  First,  $\theta$ measures the average partial effect, or ``slope,'' when characterizing the relationship between the outcome and treatment.  Second, $X^\top \gamma$ adjusts for other variables that impact both the treatment and outcome.  Third, given the observed data ($D_n$), readily available software can produce an estimate $\widehat \theta$ through the method of least squares.  Fourth, inference on the average partial effect, $\theta$, uses a confidence interval of the form
\begin{align*}
CI(\widehat \theta,\widehat \Var(\widehat \theta),\mathcal D_n,C, \alpha)=(\widehat \theta - C_{1-\alpha/2} \sqrt{\widehat \Var(\widehat \theta) },\widehat \theta + C_{1-\alpha/2} \sqrt{\widehat \Var(\widehat \theta) })
\end{align*}
where $C_{1-\alpha/2}$ is a critical value that controls the false positive rate (e.g., under a normality assumption on the error terms, we can take $1.64$ for $\alpha=0.1$ or $1.96$ for $\alpha=0.05$) given the variance of the estimated slope coefficient on the treatment variable, $\widehat \theta$. 

While this regression model is useful and versatile, these results rely on assumptions that the model makes about the relationship between the outcome, treatment, and covariates. In this paper we move past this ubiquitous implementation to a more flexible model of the relationship between outcome, treatment, and covariates.  For example, we relax the assumption that the covariates in $X$ enter linearly, and the researcher need not specify how they enter. Rather, we allow this relationship to be learned from the data. We also relax the assumption that the slope $\theta$ is homogeneous over the sample.  Instead, we allow this value to vary with the value of the treatment variable $T$ (e.g., the effect could be a curve rather than a straight line) and pre-treatment covariates $X$.  To do this, we will replace the linear component $\theta T$ with a flexible, interactive function $\theta(T,X)$. Then, we can model the effect of $T$ on $Y$ as the partial derivative of $\theta(T,X)$ with respect to $T$, denoted $\tau(T,X)$, which is the ``slope'' coefficient at a particular value of the treatment and covariates. 

However, allowing for this flexibility makes characterizing the uncertainty in our estimates (``inference'')  challenging. Adjusting for covariates nonparametrically and then allowing a flexible treatment effect falls under  a class of models referred to as {\it seminonparametric} \citep{chen:07}.  Seminonparametric models raise difficult--but surmountable--issues of how to conduct inference. On the estimation side, a crucial step is a {\it split-sample} approach, using half the sample to model the covariates and the other half to learn the structure and form of the treatment effect.  The roles of the splits are reversed and the process repeated. Doing this serves as a crucial guard against over-fitting, as one subset of the data is used to learn how the covariates enter the model and a separate set to learn the treatment effect.

However, the seminonparametric construction raises even deeper issues for inference.  While developing a confidence band for a parametric curve is straightforward, implementing one for an arbitrarily complex, nonparametric curve is rather challenging. In Section~\ref{sec:uncertainty} we develop a confidence band around $\widehat \tau$ with an {\it average coverage} property \citep{Nychka1988, wass:06}.  This band has the nice property that, for user-controlled false positive rate $0<\alpha<1$, the band will contain the true curve over at least $100 \times (1-\alpha)\%$ of the observed data, asymptotically. Because we want to consider the treatment effect as a curve, we must conduct inference on the entire curve, not simply at a single point along it. Of course, if the true model is in fact the standard linear model, then our confidence band will contain the same interval listed above (for a given $\alpha$), but only as the sample size increases.\footnote{As discussed below, our band will actually be wider than this interval, to allow for uncertainty attributable to model misspecification. Our constructed band  will contain the true value more than $100\times (1-\alpha)\%$ of the time.  Were it arbitrarily complex (which we define specifically below), the band will contain the true value $100\times (1-\alpha)\%$  of the time asymptotically.} 
As we discuss below, we turn to the method of {\it conformal inference} developed by \citet{lei:was:14}, in order to determine the critical value of the band. We show that integrating conformal inference with a machine learning approach for estimating the treatment effect and its variance results in asymptotically valid bands.

In the next section, we discuss the role of assumptions in regression models and build up to the seminonparametric approach we use. Section~\ref{sec:uncertainty} then discusses our approach to uncertainty estimation (inference). These provide the requisite pieces that then guide the MDEI algorithm presented in Section~\ref{sec:estimation}.

\section{Models and Point Estimates}
\label{sec:classes}

Starting from the simple linear regression, we progressively relax more and more assumptions until we end up with a specification in which the impact of the treatment, and both the role of the background covariates {\it and} how they (may) interact with treatment variable, is learned from the data (rather than assumed ex ante). As the models grow more complex, methods for both point estimation and inference require more nuance. We start with the simplest possible model, the linear regression, and progress to our model.

\paragraph{Existing Models}
Most published work utilizes some version of the simple regression model above: the treatment is entered linearly, is simply included additively along with additional covariates in the outcome and the treatment assignment mechanism is not modeled. This fails to capture what we are interested in modeling, which is the effect of a fluctuation of $T$ on $Y$, at some particular point $(T,X)=(T^\prime, X^\prime)$.

As a first attempt, we may choose to  maintain linearity assumptions, modeling the outcome and treatment with a regression model,
\begin{align}
\begin{split}
    Y=\theta T +X^\top \gamma +\epsilon,\;\; \E(\epsilon|T,X)=0\\
    T= X^\top \beta+u,\;\; \E(u|X)=0,\label{eq:model0}
    \end{split}
\end{align}
which we will refer to as the outcome model and treatment model, respectively.
 With standard assumptions such as no omitted confounders and no heterogeneity in the treatment effect, we can interpret $\theta$ as a causal effect; absent these assumptions, it is simply an average slope on the treatment \citep{aronow2016does}. While the treatment model in this case is not necessary, in more advanced settings, modeling the effect of a one-unit move in the treatment on the outcome will require a flexible model of the treatment variable itself.

A more challenging case emerges when the treatment connects nonlinearly with the covariates through some known function $g$:
\begin{align}
    \begin{split}
    Y=\theta T +X^\top \gamma +\epsilon,\;\; \E(\epsilon|T,X)=0\\
    T= g(X^\top \beta)+u,\;\; \E(u|X)=0.\label{eq:model1}
        \end{split}
\end{align}
Then, we can recover a consistent estimate on $\theta$ under the assumption that the model is correct: the covariates enter the outcome model linearly and enter the treatment model linearly under link $g$.\footnote{In the circumscribed case of a binary treatment $T\in\{0,1\}$  and $g()$ a logit or probit, a well-developed literature exists using matching and weighting methods. \citet{imbens2015causal} provide an overview and \citet{Sekhon2009, ho:imai:king:stua:07} provide excellent introductions for political scientists. Appendix~\ref{app:bincat} connects our approach to the binary treatment setting.} 

However, we may not believe that the covariates enter the model in a linear or additive fashion, or that we even know the function $g$.  In this case, we may turn to a model of the form
\begin{align}
    \begin{split}
    Y=\theta T +f(X) +\epsilon,\;\; \E(\epsilon|T,X)=0\\
    T= g(X)+u,\;\; \E(u|X)=0.\label{eq:model2}
    \end{split}
\end{align}
We now assume that the function $f, g$ are unspecified and must be learned from the data. This is referred to as a {\it partially linear model} \citep{Hardle89, hardle2012nonparametric}, since it is linear in $T$ but nonparametric in the remainder.

Quite a bit of work has been done in this setting, with the aim of estimating an average effect in the presence of unspecified confounding, using machine learning to adjust for the covariates.  In the case of a binary treatment, \citet{Diamond2012} introduces such a method,
but it has since been extended to continuous or arbitrary treatment variables \citep{chern:chet:2018,  kenn:lorch:inpress, fong:hazl:2018}.\footnote{Recovering a valid confidence interval on $\theta$ in this setting requires combining machine learning methods for $f,g$ with a regression for $\theta$.  A full theoretical discussion of these issues can be found in  \citet[][ch. 25]{chern:chet:2018,farreleconometrica, vand:1998}.}

The partially linear model improves on the standard practice captured in Model \ref{eq:model1}, because it allows for the fact that the confounding variables may not be linear. However, the partially linear model still assumes that the treatment enters the outcome model linearly, after adjusting for $X$ nonparametrically.  We can relax this assumption, generating a type of {\it generalized additive model} \citep{Wahba:1990, hastie1990generalized, wood:06,beck1998beyond}, where we replace $\theta T$ with $\theta(T)$,  a smooth function of $T$.  Doing so generates the model
  \begin{align}
     \begin{split}
         Y&=\theta(T)+ f(X)+\epsilon, \E(\epsilon|X,T)=0;\\
         T&=g(X)+u, \E(u|X)=0;\; f,g, \theta \textrm{ unknown}\label{eq:model3}
     \end{split}
 \end{align}
 These GAMs, also known as smoothing spline models, are used in political science and other social sciences \citep[e.g.,][]{beck1998beyond,andersen2009nonparametric,imai2010general, carter2010back,kropko_harden_2020}.
 
 If we allow $\theta$ to be a smooth function of $T$, estimation can occur following the same cross-fitting described earlier.  Rather than simply regressing $\widetilde Y$ on $\widetilde T$, we could instead model the relationship using a smoothing spline. Under the assumption that $\theta(T)$ is indeed smooth, the errors are of equal variance, and there are no treatment/covariate interactions, well-established theory and standard software can return a confidence band with average coverage (see \citet[][ch. 4]{Nychka1988} and the associated {\bf R} package {\tt mgcv} \citep{wood:06}). If the researcher is confident that these assumptions hold, this is an appropriate method.

\paragraph{A More Flexible Model}

We wish to allow for the background covariate specification to be learned from the data {\it and} the effect of $T$ on $Y$ to be nonlinear and moderated by the covariates.  We implement a model of the form
\begin{align}
    \begin{split}
    Y=\theta(T,X) +f(X) +\epsilon,\;\; \E(\epsilon|T,X)=0\\
    T= g(X)+u,\;\; \E(u|X)=0.\label{eq:model4}
        \end{split}
\end{align}
where the functions $\theta,f,g$ are all nonparametric.

We want to use this model to learn about how the treatment impacts the outcome. In the most simple regression model that we started with, this was just the marginal effect of $T$ on the outcome $Y$, which was the slope coefficient $\theta$. We want to do something similar, but in our context the slope is not a simple constant. Instead, we want $\theta$ to depend on both the value of the treatment and covariates (i.e., $\theta(T,X)$). That is, the impact of the treatment can be both nonlinear and moderated by the covariates.

In the continuous treatment case (see Appendix~\ref{app:bincat} for discussion of the binary case) our target of inference is the first partial derivative of the outcome with respect to the treatment, given background covariates: $    \tau(T,X)=\frac{\partial}{\partial T}\theta(T,X)$. This estimand captures the impact of a ceteris paribus perturbation of the treatment on the outcome.\footnote{A causal interpretation requires ignorability holds in a continuous, open ball around $T$ for each $X$ and that $\theta$ is differentiable in $T$. This function $\tau$ may be of interest in its own right even in a purely descriptive setting.} Modeling this local confounding returns an estimate of the ``instantaneous causal effect'' (ICE), a marginal effect for an observation at treatment level $T$ and covariate profile $X$ (see e.g., \citet{Stolzenberg80}). This effect can be a curve that varies across values of the treatment and can depend on values of the covariates.\footnote{See Appendix~\ref{app:bincat} for the binary treatment case.} 

\paragraph{Existing Methods That Fit This Model}  Before discussing our particular implementation, we first discuss three sets of recent methods that have been used to estimate the same model.  The first two, Bayesian Additive Regression Trees (BART) \citep{montgomery2018tree, Hill2011} and  Random Forests \citep{Hill2011, breiman2001random}, use an average of trees to predict an outcome. These methods average a group of tree models to estimate the conditional mean, $\E(Y|T,X)$, but these tree models do not allow estimation of the effect curve $\tau$.\footnote{
A tree  uses the covariates to generate a set of mutually exclusive and exhaustive indicators over the covariate space.  For an observation in each region, the estimated value is just the average outcome at of each group (''leaf''), and forests are averages over leaves; for an excellent overview of tree methods, see \citet{montgomery2018tree}.  Since the fitted value is just an average of sample means, though, it does not return a differentiable function, so it cannot estimate $\tau$--though it can be used to estimate $\tau$ if the treatment is binary, see \citet{Hill2011, greenhern}.}

Two recent methods have modeled the $\theta$ fully nonparametrically, in a way that allows us to estimate $\tau$: generalized random forests \citep[GRF, ][]{wage:athe:2017, athe:tibs:2019} and Kernel Regularized Least Squares  \citep[KRLS][]{Hainmueller13}\footnote{For additional work, see \citep{catt:farr:feng:19}, though the method does not allow for more than a few covariates.}.  KRLS fits a high-dimensional, nonparametric regression to the model, and since its estimated conditional mean is differentiable in $T$, point estimates for $\tau$ are returned. The method returns excellent point estimates across a range of settings, but as we discuss below in Section \ref{sec:uncertainty}, its confidence intervals can be misleadingly narrow.  

GRF uses a tree method, building a tree from the covariates and then running a bivariate regression of the outcome on the treatment at each leaf.  The results are aggregated over multiple trees, bootstrap samples, and randomly selected sets of covariates into a random forest.  Since, at each point in a given tree, $\widehat \tau$ is simply a regression coefficient (conditional on being in that leaf), these  can be averaged over the forest.   It can provide remarkably accurate estimation and inference on an average effect (as in Model \ref{eq:model2}).

The primary shortcoming here is that the method will miss any nonlinearities at each leaf. As it fits a simple bivariate regression, it will fail to capture nonlinearities that are missed by the regression at each leaf. For example, if the data is generated as $Y=T^2+e$ where $T, e$ are each independent and standard normal, the generalized random forest estimate $\widehat \tau(T,X)$ will completely miss the curvature.  We illustrate this below in Section \ref{sec:illustrative}.  Before describing the MDEI algorithm, which sidesteps the estimation issues described directly above, we discuss issues pertaining to uncertainty estimation.

\section{Uncertainty Estimates and Confidence Bands} 
\label{sec:uncertainty}

As discussed above, there are various options for  estimating these heterogeneous effects. Our goal is not simply estimating the treatment effect, $\tau$, but also deriving uncertainty bounds with guaranteed theoretic properties that can guide inference.  

When conducting inference on a single coefficient in a regression, we use a confidence interval that contains a segment that is  centered on the point estimate  whose width is governed by the critical value and standard error estimate.  When extending to our seminonparametric setting, we use a {\it confidence band} instead of a confidence interval.  In this case, the point estimate and standard error are each replaced by functions, $\widehat \tau(T,X)$ and $\sqrt{\widehat \Var(\widehat \tau)(T,X)}$, and we wish to connect the confidence band to the true curve, $\tau(T,X)$.

We introduce a band that allows the researcher to assess, both graphically and statistically, whether an estimated nonlinearity or interaction likely reflects a systematic relationship in the data.  The curve has the {\it average coverage property} that, asymptotically, the $100 \times (1-\alpha)\%$ curve will contain the true treatment effect at $100 \times (1-\alpha)\%$ of the observed data.   By this means, the researcher can gauge the likelihood that an uncovered nonlinearity is differentiable from zero.

We offer three advances in constructing this curve.  First, our approach uses a data-driven method to estimate a critical value.  We build on a recently proposed method, {\it conformal inference}, which can be used to generate a curve with an average coverage property on predicted values.  The method uses the observed data to generate an band on predicted values with an average coverage property.  We extend this band to cover not just predicted values, but the true conditional mean $(\theta(T,X))$ and treatment effect $(\tau(T,X))$.  Second, as we are estimating the critical value, we are not relying on a normal approximation to achieve valid coverage.   This allows our critical values to reflect the underlying distribution of the data. Third, we show that these bands, while wide at each data point, can be used to recover valid estimates of average treatment effects. As we discuss in this section and illustrate below in Section \ref{sec:illustrative}, uncertainty bands returned by existing software can be misleadingly narrow.

Next, we discuss different types of coverage that can be achieved in nonparametric settings. Then, we introduce our approach, which combines repeated cross-fitting to guard against overfitting, a model of the conditional variance to guard against model misspecification and heteroskedasticity, and a conformal approach that provides a data-driven means to select a critical value.

\subsection{Types of Coverage}

Uncertainty intervals, like confidence intervals and confidence bands, are designed to have specific coverage properties.   For example, in the linear model with a single slope coefficient, then the coverage probability is the proportion of times, over repeated samples that the interval contains the true value.\footnote{For a single parameter $\tau \in \mathcal \Re$, then, a valid $100\times (1-\alpha)\%$ confidence interval can be characterized as

$\lim_{n \rightarrow \infty} \sup_{\tau \in \mathcal \Re} \Pr(\tau \in CI(\widehat \tau,\widehat \Var(\widehat \tau),\mathcal D_n,C, \alpha)) \ge 1-\alpha$

where the standard critical values of $C=1.64$ and $C=1.96$ give the $90\%$ and $95\%$ intervals.} This is the standard confidence interval, as taught in the context of the regression and other parametric models.  

There are multiple ways to achieve coverage on a nonparametric curve. Coverage can be achieved pointwise, uniformly, and on average.  We turn to each in turn.  The first, and most commonly encountered, is the {\it pointwise confidence interval}. This is the one returned by existing software \citep[e.g.,][]{Hainmueller13, wage:athe:2017, athey2019generalized}.  In this case, at {\it any given} point $(T^\prime,X^\prime)$, the interval will cover the true value at least $(1-\alpha)\times 100\%$ of the time.\footnote{
For any given $\tau$ that can be well-approximated by the model, this interval can be characterized as:

$\textrm{{\it For any given point} }\; (T^\prime,X^\prime),\; \lim_{n \rightarrow \infty}  \Pr(\tau(T^\prime,X^\prime) \in CI(\widehat \tau(T^\prime,X^\prime),\widehat \Var(\widehat \tau(T^\prime,X^\prime),),\mathcal D_n,C, \alpha)) \ge 1-\alpha$
}

The pointwise interval does not contain information on the whole curve.  For example, from a multiple testing perspective, a $95\%$ confidence interval at every single point is not the same as a $95\%$ band over {\it all} points. It is likely too narrow, as we show below in an illustrative simulation.  Correcting the misleading nature of these curves, and allowing for more informative and honest graphical displays, is a central goal of the project.

The second type, the {\it uniform confidence band}, will contain the {\it whole} curve $(1-\alpha)\times 100\%$ of the time over repeated sampling.  This band {\it does} contain information on the whole curve, since it will contain the full curve over repeated samples.\footnote{For any $\tau$ that can be well-approximated by the method, this curve can be characterized as $\lim_{n \rightarrow \infty}  \Pr(\textrm{For all points }\; (T^\prime,X^\prime),\;  \tau(T^\prime,X^\prime) \in CI(\widehat \tau(T^\prime,X^\prime),\widehat \Var(\widehat \tau(T^\prime,X^\prime),),\mathcal D_n,C, \alpha)) \ge 1-\alpha$}
Uniform nonparameteric confidence bands have been constructed in several settings \citep[e.g.,][]{geno:wass:05,robi:vand:06}. However, in complex settings, they shrink slowly in sample size and are too wide to be usable \citep[see, e.g.,][]{wahb:83}.    

We move onto our proposed band which implements the third type of coverage, average coverage. Rather than relying on claims across repeated samples, we instead  follow \citet{Nychka1988} (see also \citet{wass:06} ch. 5.8) and consider  {\it average coverage}, which is the probability that a confidence band contains the true value over the observed sample.\footnote{ This property can be written as:\\
$\lim_{n \rightarrow \infty} \frac{1}{n} \sum_{i=1}^n \mathbf ((T_i^\prime,X_i^\prime),\;  \tau(T_i^\prime,X_i^\prime) \in CI(\widehat \tau(T_i^\prime,X_i^\prime),\widehat \Var(\widehat \tau(T_i^\prime,X_i^\prime),),\mathcal D_n,C(D_n), \alpha)) \ge 1-\alpha$} This band has the nice property that it will contain the true curve at a high percentage of the observed data. It is also narrow enough for applied work, but with provable average coverage properties.\footnote{Formal derivations of this average coverage can be found in Appendix \ref{app:variance}.}

\subsection{A Split-Sample and Repeated Cross-Fitting Approach for Valid Inference}

In order to recover valid uncertainty estimates, we implement a {\it split-sample} strategy.  This strategy proves crucial in guarding against over-fitting the data and allowing for valid inference.  The basic issue is that using the same data to learn a single model of how
the covariates affect the outcome, the treatment, and how they moderate the treatment effect is asking too much of a single data set. Machine learning methods, prone to overfitting, will return estimates with nonnegligible bias and too-narrow standard errors (as we illustrate in Section \ref{sec:illustrative}).  

The split-sample approach cleaves the problem in two: in half the data, we learn how the covariates drive the treatment and the outcome. With these models in hand, we turn to the other half of the data to learn the treatment effect model $\tau$ itself.    The split-sample approach, though, comes with clear efficiency concerns, as we are only using half the data to model the treatment effect curve $\tau$.  

In order to restore efficiency, we follow a {\it repeated cross-fitting} strategy, recently put forth  by \citet{chern:chet:2018} in the context of the partially linear model, which is our Equation \ref{eq:model2}.  
Moving to {\it repeated cross-fitting} takes two additional steps.  First, cross-fitting involves swapping the roles of the estimation and auxiliary sample, generating an additional estimate.  This process is then repeated, averaging over multiple cross-fit estimates. Our MDEI method leverages repeated cross-fitting in its implementation.

\subsection{Modeling the Standard Errors as a Guard Against Misspecification}

We make two additional moves in generating valid bands.  The first is to guard against model misspecification through modeling the conditional variance in the confidence band.  The second, which we turn to in the next section, is deriving a critical value from the data rather than through a normal approximation.  

Pointwise intervals are valid only under an assumption that the estimated model is consistent for the true curves $\theta$ and $\tau$.  This assumption is, in fact, easily violated; see Section \ref{sec:illustrative} for an example.  To add another layer of protection against requiring that we have our model correct,  we also model the conditional variance $\widehat \Var(\widehat \tau(T,X))$.  

The rationale can be found in the idea that misspecification in the conditional mean may result in systematic patterns in the residuals (see, e.g.,  \citet{king:robe:15, ratkovic2010finding}).   If the model is misspecified in some manner, we have a second chance to get our intervals correct, through using a nonparametric model of the conditional variance.

One consequence of our approach is that our intervals will, in general, be wider than the pointwise intervals returned by other methods (see Appendix~\ref{app:variance}). We could make these intervals shorter by assuming our model is properly specified, or assuming the errors are normal.\footnote{ This is what other cutting-edge methods like kernel regularized least squares \citep{Hainmueller13} and generalized random forests \citep{athey2019generalized} do.} Making these strong assumptions produces more narrow intervals, but comes at the cost of being misleadingly precise if one of the assumptions does not hold.\footnote{ As discussed by \citet{aron:16} (see also \citet{vand:1998}, esp. ch. 5), valid standard errors on regression slope parameters can be recovered with only weak distributional assumptions on the error term and outcome.} 
By modeling the error variance, we are able to recover bands that are robust to model misspecification and heteroskedasticity  (see Appendix \ref{para:errorvariance}).

\subsection{From Predictive to Bands with Average Coverage using Conformal Inference Methods}

Our first step away from pointwise intervals was to move to modeling the conditional variance as a means of guarding against model misspecification and heteroskedasticity.  Our second step is in deriving a data-driven critical value. Instead of relying on a critical value derived from a normal approximation ($C$), we instead allow it to be estimated from the data ($C(D_n)$).    We thereby generate a band that contains the true value at some controlled portion of the data that uses the empirical distribution of the data to estimate a critical value.  Both steps are needed to achieve a band with average coverage; see Section \ref{sec:illustrative} for an illustration.

The inspiration for our approach comes from recent work on conformal inference \citep{lei:was:14, lei:sell:rina:tibs:wass:19}.  Conformal inference methods give a means to produce a predictive interval, which will contain a future realization of the outcome some controlled proportion of the time, around a single point.  Importantly, it does so without making distributional assumptions about the error terms. The MDEI algorithm extends this predictive interval to one containing the true treatment effect curve, $\tau$.

\paragraph{Constructing the Interval}

We start with a predictive interval, which is an interval likely to contain a future realization of the outcome at some point, say $(T^\prime, X^\prime)$, with some controlled probability.  Constructing these intervals normally relies on distributional assumptions on the error term.\footnote{For example, in the linear model, a predictive interval around the point  $\widehat Y^\prime={X}^\top \widehat \beta$ is of the form $\widehat Y^\prime \pm \sqrt{{(X^\prime)}^\top \widehat \Var(\widehat \beta) (X^\prime) + \widehat \sigma^2}$ which is an interval relying on asymptotic normality and recovering an accurate estimate of the error variance $\widehat \sigma^2$.} The insight of the conformal approach comes from using estimated residuals to derive the width of the confidence interval and critical value on a predictive interval.  The method is entirely data-driven, and rather remarkably achieves finite-sample coverage rates on predictive intervals.\footnote{Interest in the approach is increasing in other domains of interest to social scientists \citep[e.g.,][]{cher:wuth:zhu:19,leicandes}.}

Our focus, though, is not on unobserved values of $Y$, but on a bound that will contain $\tau$ evaluated over the sample.  We use the same basic insight of the conformal method, a data driven approach for a critical value, but construct an interval around $\tau$.  We do so in two steps. First, we use the data to estimate a critical value on the bound around $\theta$ then, second, we use the same critical value to construct a bound around $\tau$.

Our goal is to derive a  bound on the approximation error, which is how far the fitted values are from the true value $(|\widehat \theta(T^\prime, X^\prime) - \theta(T^\prime, X^\prime)|)$, such that over the sample we expect the interval to contain the true value some high proportion of the time.  We do so by manipulating two other sources of error in our model. The first is the prediction error, which is how far we expect a future value to be from its prediction $(| Y^\prime-\widehat \theta(T^\prime, X^\prime)|)$.  Conformal methods give us a confidence interval built from this term.   The second is the irreducible error, which is how far a future value will be from the true value $(| Y^\prime- \theta(T^\prime, X^\prime)|)$.  Through a chain of inequalities, we show that adding one to the critical value from the predictive conformal interval will generate a valid band on the systematic component, $\theta$ (see Appendix~\ref{app:variance} and the discussion in Appendix~\ref{app:algoverview}).

%We do not want our inference to rely on, say, normal or homoskedastic errors. Instead we calculate {\it heteroskedasticity consistent}, or simply {\it robust}, standard errors.  Under this construction, we can generate valid inference without needing to specify the true distribution.

Finally, we are not interested in bounding $\theta(T,X)$ but instead $\tau(T,X)$, its partial derivative with respect to $T$. Therefore, we add a further level of approximation, such that our interval around $\widehat \theta(T,X)$ will contain $\widehat \tau(T,X)$, but, again, we are forced to rely on asymptotic argument; see Appendix \ref{app:variance}. We fully expect the constructed confidence interval to be conservative, meaning the $100\times(1-\alpha)\%$ confidence interval will contain the true systematic component at {\it more than} $100\times(1-\alpha)\%$ of the data, even asymptotically. This is because our method is not exact and is constructed through a set of bounds on an asymptotically exact bound. That is, the predictive conformal band of \citet{lei:was:14} will hold exactly, in-sample.  Our bound takes two steps past this: first, to the systematic value $\theta$, and then to its partial derivative, $\tau$.  In doing so, we move from modeling something that is theoretically observable, a future value of $Y$, to two functions that are not, $\theta, \tau$. Generating estimates of $\theta$, $\tau$, and these intervals pose a real challenge, to which we turn next.

\section{The MDEI Algorithm}
\label{sec:estimation}

We are now ready to present how we actually estimate and conduct inference on our model, all of which is implemented in open source software. Recall that our model takes the form
\begin{align}
    \begin{split}
    Y=\theta(T,X) +f(X) +\epsilon,\;\; \E(\epsilon|T,X)=0\\
    T= g(X)+u,\;\; \E(u|X)=0.\label{eq:model5}
        \end{split}
\end{align}

In estimating and conducting inference on $\tau = \frac{\partial}{\partial T}\theta(T,X)$, we have three core goals: to allow the instantaneous effect of the treatment to be a nonparametric function of the treatment and interactions with covariates, to adjust for confounders nonparametrically,  and to recover uncertainty estimates that allow for inference. We here sketch out the primary pieces. Appendix~\ref{app:algoverview} provides a detailed step through of the MDEI algorithm. The process involves two distinct stages: a full sample stage and a split-sample stage. The intuition of having two stages is that the full sample stage--where we use all observations--will help to roughly learn the relationships between the outcome and the treatment/covariates. The split-sample stage, where we split the data in half, is used for making specific estimates of parameters and uncertainty around them.

\subsection{Full Sample Stage}

Using the full set of data, we first want to want to identify a large number of nonlinear transformations of $X$ and $T$ that might help model $\tau$ and hence $\theta$. This gives us a model space, rather than committing to a specific model. These transformations are then used in the split-sample stage where we obtain point and uncertainty estimates. 

Prior to beginning this step, we partial out individual covariates. We do so by estimating two variables,
\begin{align*}
 \widetilde Y = Y-\E(Y|X); \;\;  \widetilde T = T - \E(T|X)
\end{align*}

By this means, we eliminate any systematic relationship  $\widetilde Y$ and $\widetilde T$ that is only a function of the background covariates, i.e. it will involve the treatment variable on its own or moderated by a covariate. Following other work \citep{chern:chet:2018}, we do this via random forests so as to make extremely minimal  assumptions about the relationship between the covariates and the outcome/treatment variables (see Appendix~\ref{app:functionclassappendix}).

Next we model the treatment effect with nonparametric basis functions, also known as ``blending functions,'' that can be added together to approximate other functions.\footnote{See \citet{keele2008semiparametric} for a helpful introduction for social scientists.} These capture any complicated interactions between the treatment and covariates, as well as allow the impact of the treatment on its own to be nonlinear. Appendix~\ref{app:basisappendix} provides an introduction to basis functions and our strategy for modeling these.

For an overview, imagine we knew, or suspected, that the outcome had a quadratic nonlinearity in the treatment, so we could simply take the model
\begin{align*}
\theta(T,X) = \widetilde T c_1+ \widetilde T^2 c_2 
\end{align*}
where $c_1, c_2$ are slope parameters on the linear and quadratic term of the partialed-out treatment, $\widetilde T$.  We could then take the partial derivative to recover $\tau(T,X) = c_1 + 2 \widetilde T  c_2$. We of course do not want to restrict ourselves to the consideration of one particular type of nonlinearity (see  Appendix \ref{app:functionclassappendix} for the minimal functional form assumptions needed for consistency in our estimates) and instead let data rather than model assumptions drive our inferences.

From this additive nonlinear model, we take three additional steps.  First, we allow for nonlinear functions of the treatment.  We will denote these as $\phi_j(\widetilde T)$, where each $\phi_j$ is some transformation, indexed by $j$, of the covariate. 
We select a set of basis function transformations which are known to approximate a broad class of possible functions; Appendix~\ref{app:basisappendix} presents details.

If we knew which nonlinear transformation were driving the treatment effect (e.g., quadratic as used above), we could model
\begin{align*}
\theta(T,X) =  \phi_j(\widetilde T) c_{j}
\end{align*}
with $c_{j}$ the slope parameter.   Taking a partial derivative gives
\begin{align*}
    \tau(T,X) = \dot \phi_j(\widetilde T) c_{j}
\end{align*}
 where 
 \begin{align*}
     \dot \phi_j = \partial \phi_j/\partial \widetilde T.
 \end{align*}

Second, we may want the covariates to moderate this effect.  We could allow for a single moderator, the $j^{th}$ transformation of variable $k$, denoted $\phi_j(X_k)$, and take the model
\begin{align*}
\theta(T,X) = \phi_j(\widetilde T) \phi_{j^\prime}(X_{k^\prime})c_{jj^\prime k^\prime};\;\;
\tau(T,X) =  \dot \phi_j(\widetilde T) \phi_{j^\prime}(X_{k^\prime})c_{jj^\prime k^\prime}.
\end{align*}

We in fact consider all possible $treatment \times covariate \times covariate$ interactions.\footnote{In this case of allowing two moderators, we have:
\begin{align*}
\theta(T,X) = \phi_j(\widetilde T) \phi_{j^\prime}(X_{k^\prime})\phi_{j^{\prime\prime}}(X_{k^{\prime\prime}})c_{jj^\prime k^\prime j^{\prime\prime}k^{\prime\prime}};\;\;
\tau(T,X) =  \dot  \phi_j(\widetilde T) \phi_{j^\prime}(X_{k^\prime})\phi_{j^{\prime\prime}}(X_{k^{\prime\prime}})c_{jj^\prime k^\prime j^{\prime\prime}k^{\prime\prime}}.
\end{align*}
 } These equations capture the idea that we are modeling the treatment effect as a nonlinear function of the treatment interacted with up to two nonlinear interactions of the covariates.

The problem, though, is that we do not know which nonlinear transformations of the treatment and moderators may be driving $\theta$ and, hence, $\tau$. Instead, we consider a large set of them. We provide details in Appendix \ref{app:implementedmodel}, but for $5$ covariates, we consider over $174,000$ possible nonlinear terms and interactions and for $10$ covariates, over $615,000$.  The full set of these terms allow us to fit a vast number of possible interactions and nonlinearities.

Even with a handful of covariates, the proliferation of candidate basis functions puts us in an extremely high-dimensional space. We reduce this dimensionality in two stages: 1) an iterated Sure Independence Screen \citep{Fan:2008} and 2) a variant of the high-dimensional regression LassoPLUS model \citep{RatkovicTingley17} that we tune for the present context (see Appendix~\ref{app:sparseappendix} for details). These steps provide us with a set of candidate basis functions from which a smaller subset will be selected during the split-sample estimation stage. 

\subsection{Split-sample stage}

Armed with a subset of a vast set of candidate basis function transformations learned from the full sample of data, we next split the data into an \textit{auxiliary} and \textit{estimation} sample. The split-sample strategy helps us guard against data dredging by learning a model in the auxiliary sample but then using the estimation sample to conduct inference \citep{vand:1998,AtheyImbens16,chern:chet:2018}.  This split-sample approach is a crucial step in generating proper confidence bands: any overfitting of the machine learning in the auxiliary set will not pollute our inference in the estimation set (see \citep{chern:chet:2018} for a recent overview).  

\paragraph{Auxiliary sample} The auxiliary sample uses the candidate basis functions from the full-sample step in order to learn a plausible specification for $\theta$. As in the full sample step, we partial out the covariates in the auxiliary sample and select bases using the high-dimensional regression model. This provides two things that will be used in the the estimation sample 1) the random forest model learned from the auxiliary sample and 2) the selected bases. 

\paragraph{Estimation sample} The estimation sample is then used to generate $\widehat \theta$, $\widehat \tau$, and the confidence band around $\tau$. We first partial out the covariates given the random forest model learned on the auxiliary data. Then we estimate a regression to obtain coefficients on the selected bases that let us obtain point estimates for $\widehat \theta$ and $\widehat \tau$. We are not just interested in predicting point estimates, but also in getting confidence bands correct.  We therefore also use the auxiliary sample to predict the variance at each point, given the value of the treatment and covariates.  The square root of this variance is then used to construct our band, at each point.  We next use the conformal inference method discussed above to recover a data driven critical value that does not rely on assumptions on our error term (see Appendix~\ref{app:algoverview} and ~\ref{app:variance} for details). In doing so, we extend the work of \citet{lei:was:14} from finding a predictive confidence band to constructing a band on our 
marginal effect curve $\tau(T,X)$.

We then extend our split-sample approach to a {\it repeated cross-fitting} strategy.  This involves two steps: a cross-fitting step that is then repeated.  In the cross-fitting step,  we alternate the role of the auxiliary and estimation sets, generating two estimates of $\theta$ and $\tau$ for each split of the data.  We then repeat this process of randomly splitting the data and cross-fitting multiple times, averaging over our results; see Appendix \ref{app:algoverview} for details.

\subsection{Discussion} MDEI is able to recover a nonparametric treatment effect with point estimates competitive with several cutting-edge methods (see below and Appendix~\ref{app:performsims}). Furthermore, we provide a principled approach to uncertainty estimation. We are able to recover a confidence band that contains the true value over at least $100\times(1-\alpha)\%$ of the data points, asymptotically. In generating a curve that is valid under minimal assumptions on the functional form and error distribution, we expect it to be conservative. Decreasing the width of these bands could be done, but at the cost of additional assumptions on the data. We show in our applied example below that these bands are still narrow enough to be used for inference in applied work. In Appendices \ref{app:algoverview}--\ref{app:variance} we provide extensive technical details for interested readers.

\section{Illustrative Simulation}
\label{sec:illustrative}

Next we illustrate the role of repeated cross-fitting and conformal inference in achieving average coverage. We also compare MDEI intervals to those returned by to two cutting edge alternative algorithms.  We show that MDEI performs well in terms of point estimation and, equally importantly, returns intervals that relay information on where the true systematic curve is likely to lie.\footnote{ Appendix~\ref{app:performsims} provides a complete set of Monte Carlo performance simulations. }

%As a reminder, our goal is to generate a method that can allow us to estimate how a  treatment influences some outcome, after adjusting for pre-treatment covariates that might confound the relationship between the treatment and outcome. We want to estimate and, subsequently, conduct inference on a marginal effect curve. This marginal effect curve could be non-linear, and it could depend on the value of other covariates. [dt put roadmap when done]

\paragraph{Illustrating the role of split-sample/repeated cross-fitting and conformal approach}
In the first set of illustrative simulations, we show how repeated cross-fitting and the conformal approach can lead to valid average coverage. Our simulation setting has three important characteristics.  First, we include a nonlinear confounder that must be learned from the data.  Second, the treatment effect is moderated by one of the covariates in a discontinuous fashion, making it a challenge for our method to uncover.  Third, we include heteroskedastic errors in order to illustrate the method's coverage even with systematic components in the error variance.
%Third, we vary whether the errors are equivariant or whether their variance depends on the moderating covariate, again in a nonlinear fashion.

In each setting,  we draw five covariates from a standard multivariate normal equicorrelated at $0.5$. From the first,  $X_{i1}$, we generate a new variable $s_i=\sgn(X_{i1})\in \pm 1$.  This sign function, a discontinuous function of a continuous covariate, will serve to govern the effect heterogeneity: the impact of the treatment on the outcome will vary with whether this first variable is positive or negative.   The outcome and the treatment are generated as:
  \begin{align}
       T_i &= \frac{(X_{i2}-1)^2}{4}+u_i;\;u_i \sim \mathcal N(0,1)\\
    Y_i &= 2 s_i T_i^2+\frac{(X_{i2}-1)^2}{4} + \epsilon_i;\; \epsilon_i \sim \mathcal N\left(0,\frac{1}{1+X_{i2}^2}\right).
 \end{align}
 This gives us a target of $\tau(T_i,X_i) = 4s_iT_i$
for which we want to use our interval to conduct inference. We vary the sample size, $n \in$ 100,250,500,1000,2500,5000,10000 and simulations were run $400$ times each.

\begin{figure}[t!]
%\vspace{-.9in}
\begin{center}\spacingset{1}
\includegraphics[scale=.6]{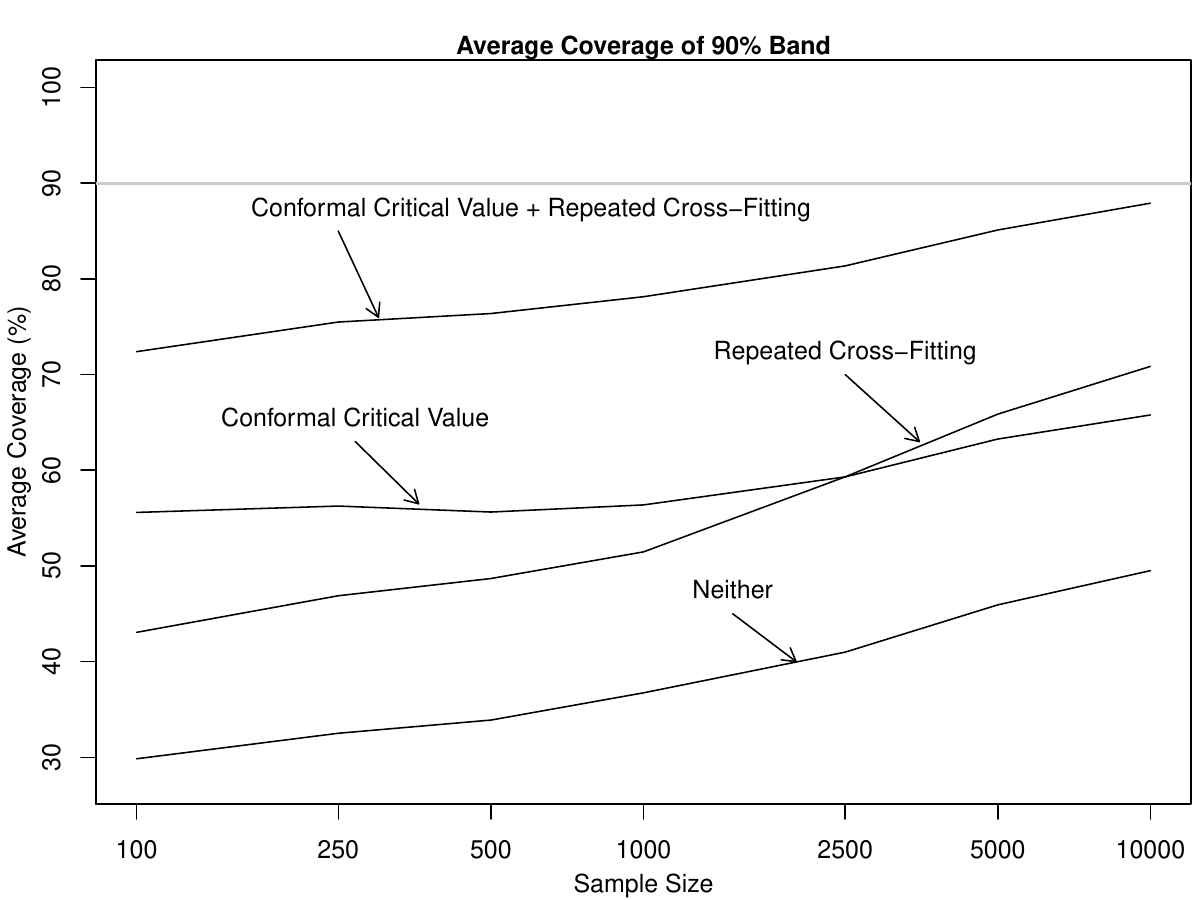}
\vspace{-.25in}
\end{center}\spacingset{1}
\caption{{\bf Average coverage of 90\% band} 
Each lines corresponds with the four possible settings for our confidence bands. The  pointwise interval will be quite narrow, not actually containing the true curve in the majority of the data.  The conformal critical value increases coverage, but only when combined with repeated cross-fitting does coverage approach nominal.
} \label{fig:illussim1}
\end{figure}

Results appear in Figure~\ref{fig:illussim1}.  For each run of the simulation, we calculate a $90\%$ interval, and assess what proportion of the data is contained in the constructed interval. The $y$-axis presents average coverage, with sample size on the $x$-axis.  The solid horizontal line at 90\% is the expected coverage. Each of the four lines corresponds with the four possible settings for how we construct our confidence bands: with neither a conformal critical value nor repeated cross-fitting; with either repeated cross-fitting or a conformal critical value of $1.64$; and both the conformal and repeated cross-fitting strategy. 

%Long version pre JOP Each lines corresponds with the four possible settings for our confidence bands based off whether we use repeated cross-fitting and whether we use a conformal critical value or the standard critical value for a $90\%$ pointwise interval of $1.64$. The  pointwise interval will be quite narrow, not actually containing the true curve in the majority of the data.  The conformal critical value increases coverage, but only when combined with repeated cross-fitting does coverage approach nominal. The result is clearly asymptotic, but even in reasonable sample sizes, our 90\% band achieves coverage of $>80\%$ in this complex setting, suggesting our method can be used in settings with realistic sample sizes. 

The figure shows that the pointwise interval will be quite narrow, not actually containing the true curve in the majority of the data.  The conformal critical value serves to recover bands closer to the expected average coverage, but only when combined with repeated cross-fitting do we find coverage close to what is expected. The result is clearly asymptotic, but even in small sample sizes, our 90\% band achieves coverage of $>80\%$ in this complex setting, suggesting our method can be used in settings with realistic sample sizes.

\paragraph{Illustrating the difference between pointwise and average coverage}

Next, we compare the pointwise interval returned by existing methods to those with average coverage, returned by MDEI.  We show that the pointwise intervals can be so narrow as to generate misleading inference. Using the same data generation process as before and a sample size of $5,000$, we focus on the contrast between MDEI\footnote{Estimation with MDEI is done in {\bf R} through one line of code, s1 $\leftarrow$ sparseregTE(Y=y, treat=treat, X=X) where $X$ is simply a matrix of pre-treatment covariates. No additional inputs are required from the user.} and both Kernel Regularized Least Squares ($KRLS$, \citep{Hainmueller13}) and Generalized Random Forests ($GRF$, \citep{athey2019generalized}). Both can estimate the marginal effect curve directly.

 \begin{figure}[t!]
%\vspace{-.9in}
\begin{center}\spacingset{1}
\includegraphics[scale=.8]{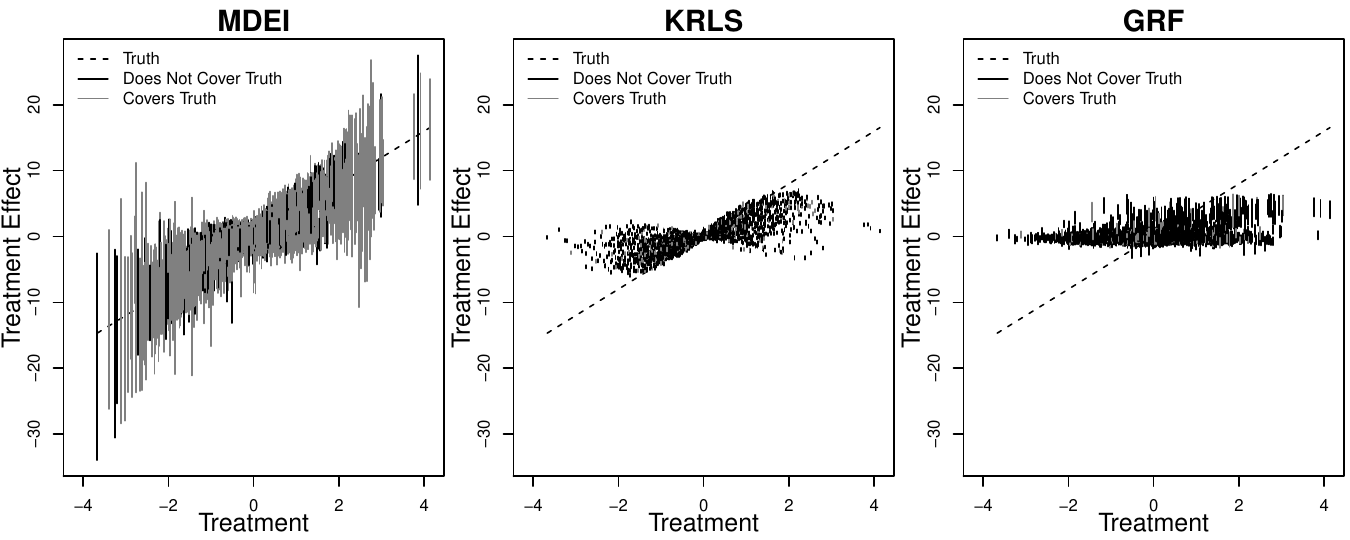}
\vspace{-.25in}
\end{center}\spacingset{1}
\caption{{\bf Treatment effect estimates across MDEI, KRLS and GRF algorithms.} The dashed line represents the true treatment effect, with results by method in each plot.  Vertical lines represent the 90\% uncertainty intervals from each method. Only MDEI is able to capture the trend in the treatment effect, while the uncertainty estimates from other methods are narrow. } \label{fig:illussim2}
\end{figure}

%pre JOP The dashed line represents the true treatment effect, with results by method in each plot.  Vertical lines represent the 90\% uncertainty intervals from each method. Only MDEI is able to capture the trend in the treatment effect, while the uncertainty estimates from other methods are narrow and cover the true treatment effect curve only incidentally.

\begin{comment}
%>     plot.te(2,"MDEI")
%   -1 
0.846
>     plot.te(3,"KRLS")
    -1 
0.1468 
>     plot.te(4,"GRF")
    -1 
0.1218 
    
>     colMeans(abs(m1$mce.true-te))
   truth     MDEI     KRLS      GRF 
0.000000 1.348685 1.985816 3.312850 
\end{comment}

Figure~\ref{fig:illussim2} displays the results, plotting the treatment effect curve ($\tau$) against its estimate and interval, by method.  We present results for $s_i=1$; they are qualitatively similar to $s_i= -1$.   Intervals that contain the truth are in grey, and those that are not are in black.  The 90\% band returned MDEI covers the true value at $84.6\%$ of the data, in this example.  For KRLS and GRF, those numbers are $14.7\%$ and $12.2\%$.  The root-mean square error across the methods reveal a similar pattern (MDEI: 1.34, KRLS: 1.98, GRF: 3.31).  GRF uses a split sample approach, while KRLS does not, but still achieves a higher error and lower coverage because, as mentioned above, GRF cannot capture curvature in the treatment variable well.  MDEI, on the other hand, returns point estimates and an uncertainty band that convey the basic nonlinearity in the underlying data as well as containing the true curve about as frequently as would be expected.

\section{Applied Example}\label{sec:example}

\citet{bech:hain:2011} explore the impact of an effective policy response to a natural disaster in Germany.  They estimate the effect of the government's successful response to the 2002 flooding of the Elbe River on support for the incumbent party, the Social Democrats, in the 2002 federal elections.  Using a difference-in-difference design with a linear regression specification, the authors estimate an impact of approximately 7 percentage points on the Social Democrats' vote share.

Applying MDEI to the same data with the same control set, we find a  smaller--while still statistically significant--effect.  Here, the outcome is the change in voteshare for the Social Democrats, the treatment variable is whether the region was flooded, and the controls include a battery of covariates that adjust for sociodemographic and economic factors \citep[see][Table 1, pg. 857]{bech:hain:2011}.  

We start by analyzing this situation, with a binary treatment variable (was a district flooded or not?) in order to build faith in the method (see Appendix~\ref{app:bincat} for estimation details in this binary setting).  Estimates of the effect on change in vote using Bechtel and Hainmueller's data  appear in Table \ref{tab:bh}.   The first row contains the results using the control set in the original work.  The results from MDEI appear in the second line, using the same control set, outcome, and treatment.  To calculate the effect, we took the average effect on all flooded districts, averaged their variances, and used the critical value returned by the method.  We find a point estimate lower than the original analysis, though still significant.  We find that the discrepancy between the original results and MDEI is likely due to covariate imbalance between treated and untreated regions.  If we trim districts further from the Elbe than the treated districts and then run Bechtel and Hainmueller's specification, we recover an estimate that is much closer to that from  MDEI. For further verification, we compare the results to generalized random forests (GRF) as well as the authors' original specification on the trimmed data, but using a smoothing spline in distance from the Elbe (GAM).  We see that all of the methods besides the original regression agree on the magnitude of the effect, with MDEI providing the narrowest confidence interval.

The reason for the improved performance is implicit in the method. The estimand for the difference-in-difference design is the average effect on the treated districts.  The difference-in-difference regression coefficient, though, is  a weighted average of the difference between treated and untreated units. If the untreated units are not directly comparable with the treated units, the coefficient may be biased.  This is what we see here.  In contrast, MDEI returns an estimated effect for each point, and then we aggregate only over the treated units in order to estimate the treatment effect on the treated.  Doing so reduces concerns over imbalance, as we are only using observations that were in fact treated when estimating the treatment effect on the treated.
\begin{comment}
 estimate   std.err 
4.5691673 0.4609759 
\end{comment}

\begin{table}[t!]
\spacingset{1}
\centering
\begin{tabular}{rrc}
  \hline
 & Point Estimate & 95\% CI \\ 
  \hline
Original Regression & 6.91 & (5.43, 8.40) \\ 
  Trimmed Regression & 4.87 &(2.96, 6.78) \\ 
    MDEI &4.61 & (4.19,  5.03) \\ 
    GRF & 4.56 & (3.65, 5.46)\\
    GAM & 4.55 & (3.27, 5.82)\\
   \hline
\end{tabular} \spacingset{1}
\caption{{\bf Estimates Across Model Specifications.} Rows contain the estimated effect on the Social Democrat's voteshare in flooded regions from the original specification, a trimmed regression, $GRF$, and a $GAM$ using a trimmed regression but adding a smoothing spline in distance from the river. 
}\label{tab:bh}
\end{table}

%pre jop version Rows contain the estimated effect on the Social Democrat's voteshare in flooded regions from the original specification, MDEI, and trimming unflooded districts further than the furthest flooded district, respectively.  MDEI returns a point estimate that is lower than the original regression, but much closer to that from a regression on data trimming observations further from the Elbe than the furthest flooded region.  It also agrees with the effect returned by random forest ($GRF$) as well as using a trimmed regression but adding a smoothing spline in distance from the river ($GAM$).  The confidence intervals on the average effect from MDEI are also narrower than those from the other methods, suggesting that it allows for accurate and efficient inference.

\begin{figure}[t!]
%\vspace{-.9in}
\begin{center}\spacingset{1}
\includegraphics[scale=.75]{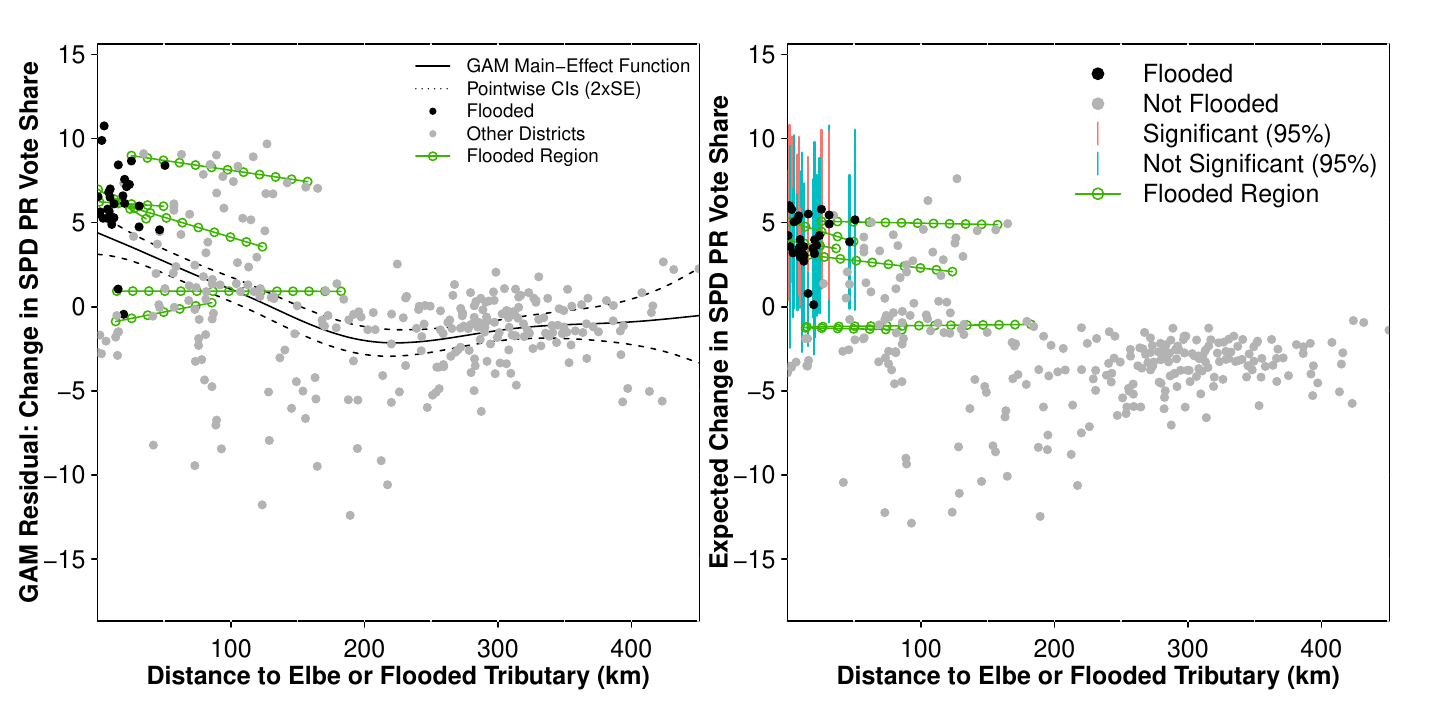}
\vspace{-.25in}
\end{center}\spacingset{1}
\caption{{\bf The Effect of Distance to Elbe on Vote Share.} 
The left hand figure presents the estimated effect of distance to the Elbe on voteshare using the specification from Bechtel and Hainmueller.  Analyzing residuals,  flooded districts (solid black dots) are systematically above the trend.  Regions with flooding exhibit a negative trend, suggesting that  the effect is due to flooding and not some other confounding  event.  MDEI, on the right, is able to recover similar results as in the original paper, but in one step and with uncertainty bands.} \label{fig:bhdistance}
\end{figure}

We next estimate the effect of a continuous treatment on an outcome.  Bechtel and Hainmueller argue that the effect of policy response on voteshare decays as the distance from the Elbe increases for regions in which there were flooded districts, which they argue is further evidence that the discovered effect is attributable to disaster response.  We reevaluate both claims and present results in Figure \ref{fig:bhdistance}. We begin with their analysis (see Figure 5 in the original paper),\footnote{We combine both halves of their Figure 5 into one plot for parsimony.} which we present in the left-hand panel of Figure \ref{fig:bhdistance}.  The authors fit a smoothing spline (GAM), smooth in distance to the Elbe with the same set of linear controls included as before.  Examining the residuals,  flooded districts (solid black dots) are systematically above the trend, suggesting that these observations are systematically high.  Then, the authors fit lines to the residuals by region containing districts that flooded, which we present as the green lines. The slopes of four of these lines are negative, which they argue suggests the effect is due to flooding and not some other confounding variable or concurrent political event.

%Pre JOP The lefthand figure presents the estimated effect of distance to the Elbe on voteshare using the specification from Bechtel and Hainmueller.  Fitting the model and then analyzing the residuals,  flooded districts (solid black dots) are systematically above the trend.  Trends in regions with flooding are in green, and the authors use the negative slopes to argue that  the impact on voteshare is due to flooding and not some other confounding  event.  The righthand figure gives the analysis from MDEI, where we can directly examine fitted values from the model.  The 95\% confidence band at flooded districts is, for the most part, above zero, and we find the same downward trends in flooded regions.  We are able to recover similar results as in the original paper, but in one step, and with uncertainty bands. 

In the right panel, we present the fitted values from MDEI where we take each district's distance to the Elbe as the treatment.  We include the authors' original covariates and add in controls for whether the district flooded and whether the district is in a region that had at least one flooded district.  We find similar trends in regions where districts were flooded, but we find them in the fitted values rather than the residuals.  The original work  analyzed residuals, after taking out a smooth trend in distance and additive covariates. The righthand panel, using MDEI, uncovers the same effects through the model and the covariates.

Exploring the fitted values is preferable, because we can attribute their values to observed covariates, as compared to estimating with residuals which are, by design, noisy. It also allows us to estimate and analyze effects in one step, looking at fitted values and bands, rather than the two- step process of estimating fitted values and looking at the residual.  Using our method, we find a similar pattern: with flooded districts are systematically above zero, meaning the vote share for the Social Democrats went up, and there is less variance in the segments fit to regions where there was flooding than to unflooded regions.

\begin{figure}[t!]
%\vspace{-.9in}
\begin{center}\spacingset{1}
\includegraphics[scale=.75]{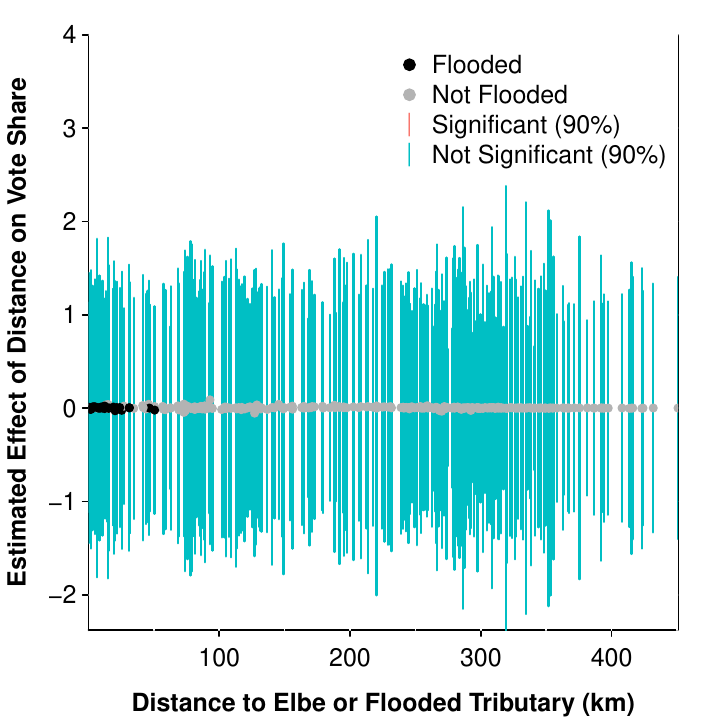}
\vspace{-.25in}
\end{center}\spacingset{1}
\caption{{\bf The Effect of Distance to Elbe on Vote Share.} The point estimate and confidence interval of the marginal effect on distance to the Elbe on support for Social Democrats, by observation.  Point estimates for districts in flooded regions are in black, the rest gray.   Distance has no discernible effect on support for Social Democrats.
  } \label{fig:bhslope}
\end{figure}

Although we find a similar pattern in the data as Bechtel and Hainmueller, we now want to know  whether it is distance from the Elbe, or simply having been flooded, that is driving the estimated effect on voteshare.  Figure \ref{fig:bhslope} presents the estimated effect of distance on voteshare at each point, with flooded districts black and non-flooded grey.  After adjusting for the other covariates, we find no effect at any observation.  Our analysis seems to suggest that the relationship between distance to the Elbe and voteshare is null, after adjusting for flooding and other covariates.  The estimated effect seems attributable to whether the district was flooded, and not to its distance from the Elbe.

Our reanalysis has recovered the central finding of Bechtel and Hainmueller, that flooded districts rewarded the Social Democrats.  At the same time, we found the effect to be somewhat overstated likely due to the inclusion of non-flooded districts that were not directly comparable to the flooded districts. We then found evidence that the result is being driven by whether a district is flooded, and not its distance from the Elbe. Throughout we are able to entertain non-linear effects as well as recover uncertainty estimates.

\section{Conclusion}
\label{sec:conclusion}

A central challenge in regression analysis is correctly modeling how a treatment variable impacts an outcome. Is the effect non-linear? Does it depend on the values of other variables, or a combination of both? Traditional regression models grow increasingly unhelpful given these challenges, especially as the number of variables and potential non-linear relationships increases.  We introduce an estimation process that allows for the seminonparametric estimation of a treatment effect \textit{and} robust uncertainty estimates. 

%Beyond the concrete methodological contribution, we also provide a heuristic understanding of the challenges of both estimation and inference when we want to relax standard modelling assumptions. 

We hone in on the type of inference that is appropriate when estimating nonlinear relationships when we do not ex ante specify specific nonlinear relationships. The method we propose builds on recent work involving iterated cross-fitting and conformal inference. Simulation evidence shows that the proposed method performs very well.

While we dramatically reduce reliance on ex ante modelling choices, we do of course retain other assumptions required for making causal claims (e.g., no omitted confounders). The approach presented in this paper also does not deal with other challenges to causal inference (e.g., improper confounding strategies such as controlling for post-treatment variables or certain types of pre-treatment variables \citep{acharya2015explaining, morgan2014counterfactuals,glynn2018front}, which are research questions that precede the choice of model. In a separate paper we discuss how to extend our framework to the instrumental variables and causal mediation frameworks. 

\spacingset{1.58}
%\clearpage
\bibliography{bibliography}

@book{hastie1990generalized,
  author = {Hastie, Trevor and Tibshirani, Robert},
  publisher = {Wiley Online Library},
  title = {Generalized additive models},
  year = 1990
}

@Unpublished{lei:sell:rina:tibs:wass:19,
  author = {Jing Lei and Max G'Sell and Alessandro Rinaldo and Ryan J. Tibshirani and Larry Wasserman},
  title  = {Distribution-Free Predictive Inference for Regression},
  note   = {arXiv:1604.04173v2},
  month  = mar,
  year   = {2017},
}

@Unpublished{cher:wuth:zhu:19,
  author = {Victor Chernozhukov and Kaspar Wuthrich and Yinchu Zhu},
  title  = {An Exact and Robust Conformal Inference Method for Counterfactual and Synthetic Controls},
  note   = {arXiv:1712.09089},
  year   = {working},
}

@Article{farreleconometrica,
  author  = {Max Farrell and Tengyuan Liang and Sanjog Misra},
  title   = {Deep Neural Networkds for Estimation and Inference},
  journal = {Econometrica},
  year    = {forthcoming},
  volume  = {},
  number  = {},
  pages   = {},
}

@Article{robi:vand:06,
  author  = {James Robins and Aad van der Vaart},
  title   = {Adaptive Nonparametric Confidence Sets},
  journal = {Annals of Statistics},
  year    = {2006},
  volume  = {34},
  number  = {1},
  pages   = {229--253},
}

@Article{wahb:83,
  author  = {Grace Wahba},
  title   = {Bayesian ``Confidence Intervals'' for the Cross-Validated Smoothing Spline},
  journal = {Journal of the Royal Statirtical Society, Ser. B},
  year    = {1983},
  volume  = {45},
  pages   = {133-150},
}

@Article{geno:wass:05,
  author  = {Christopher R. Genovese and Larry Wasserman},
  title   = {Confidence sets for nonparametric wavelet regression.},
  journal = {Annals of Statistics},
  year    = {2005},
  volume  = {33},
  number  = {2},
  pages   = {698--729},
}

@Unpublished{aron:16,
  author = {Peter Aronow},
  title  = {A Note on "How Robust Standard Errors Expose Methodological Problems They Do Not Fix, and What to Do About It."},
  note   = {arXiv Manuscript},
  month  = sep,
  year   = {2016},
}

@Article{king:robe:15,
  author  = {Gary King and Margaret E. Roberts},
  title   = {How Robust Standard Errors Expose Methodological Problems They Do Not Fix, and What to Do About It.},
  journal = {Political Analysis},
  year    = {2015},
  volume  = {23},
  number  = {2},
  pages   = {159--178},
}

@Book{wood:06,
  title     = {Generalized Additive Models: An Introduction with R},
  publisher = {Chapman \& Hall/CRC Texts in Statistical Science},
  year      = {2006},
  author    = {Simon N. Wood},
}

@Article{lei:was:14,
  author  = {Jing Lei and Larry Wasserman},
  title   = {Distribution-free prediction bands for nonparametric regression},
  journal = {Journal of the Royal Statistical Society (Series B)},
  year    = {2014},
  volume  = {76},
  number  = {1},
  pages   = {71--96},
}

@article{Nychka1988,
	Author = {Nychka, Douglas},
	Confidential = {n},
	Journal = {Journal of the American Statistical Association},
	Pages = {1134-1143},
	Title = {Bayesian Confidence Intervals for Smoothing Splines},
	Volume = {83},
	Year = {1988}}

@Book{wass:06,
  title     = {All of Nonparametric Statistics},
  publisher = {Springer},
  year      = {2006},
  author    = {Larry Wasserman},
  series    = {Springer Texts in Statistics},
}

@Article{catt:farr:feng:19,
  author  = {Matias D. Cattaneo and Max H. Farrell and Yingjie Feng},
  title   = {Large Sample Properties of Partitioning-Based Series Estimators},
  journal = {Annals of Statistics},
  year    = {Forthcoming},
}

@article{glynn2018front,
  title={Front-door versus back-door adjustment with unmeasured confounding: Bias Formulas for front-door and hybrid adjustments with application to a job training program},
  author={Glynn, Adam N and Kashin, Konstantin},
  journal={Journal of the American Statistical Association},
  volume={113},
  number={523},
  pages={1040--1049},
  year={2018},
  publisher={Taylor \& Francis}
}

@article{athey2019generalized,
  title={Generalized random forests},
  author={Athey, Susan and Tibshirani, Julie and Wager, Stefan and others},
  journal={The Annals of Statistics},
  volume={47},
  number={2},
  pages={1148--1178},
  year={2019},
  publisher={Institute of Mathematical Statistics}
}

@InBook{chen:07,
  chapter   = {Large Sample Sieve Estimation of Semi-Nonparametric Model},
  pages     = {5549--5632},
  title     = {Handbook of Econometrics},
  publisher = {Elsevier},
  year      = {2007},
  author    = {Xiaohong Chen},
  volume    = {6 Part B},
}

@article{Stolzenberg80,
	Author = {Ross Stolzenberg},
	Journal = {Sociological Methodology},
	Pages = {459--488},
	Title = {The Measurement and Decomposition of Causal Effects in Nonlinear and Nonadditive Models},
	Volume = {11},
	Year = {1980}}

@article{Belloni14,
	Author = {Alexandre Belloni and Victor Chernozhukov and Christian Hansen},
	Journal = {Review of Economic Studies},
	Number = {2},
	Pages = {608--650},
	Title = {Inference on Treatment Effects after Selection among High-Dimensional Controls},
	Volume = {81},
	Year = {2014}}

@article{holl:86,
	Author = {Holland, Paul W.},
	Journal = {Journal of the American Statistical Association},
	Pages = {945--960},
	Title = {Statistics and Causal Inference (with Discussion).},
	Volume = 81,
	Year = 1986}

@article{Hardle89,
	Author = {Wolfgang Hardle and Thomas M. Stoker},
	Journal = {Journal of American Statistical Association},
	Pages = {986--95},
	Title = {Investigating Smooth Multiple Regression by the Method of Average Derivatives},
	Volume = {84},
	Year = {1989}}

@book{hardle2012nonparametric,
	Author = {H{\"a}rdle, Wolfgang Karl and M{\"u}ller, Marlene and Sperlich, Stefan and Werwatz, Axel},
	Publisher = {Springer Science \& Business Media},
	Title = {Nonparametric and semiparametric models},
	Year = {2012}}

@article{Ridgeway99,
	Author = {Greg Ridgeway},
	Journal = {Computing Science and Statistics},
	Pages = {172--181},
	Title = {The state of boosting},
	Volume = {31},
	Year = {1999}}

@article{Ravikumar09,
	Author = {Pradeep Ravikumar and John Lafferty and Han Liu and Larry Wasserman},
	Journal = {Journal of the Royal Statistcal Society, Series B},
	Number = {5},
	Pages = {1009--1030},
	Title = {Sparse Additive Models},
	Volume = {71},
	Year = {2009}}

@article{ho:imai:king:stua:07,
	Author = {Ho, Daniel E. and Imai, Kosuke and King, Gary and Stuart, Elizabeth A.},
	Journal = {Political Analysis},
	Month = {Summer},
	Number = {3},
	Pages = {199--236},
	Title = {Matching as Nonparametric Preprocessing for Reducing Model Dependence in Parametric Causal Inference},
	Volume = {15},
	Year = {2007}}

@article{Diamond2012,
	Author = {Diamond, Alexis and Sekhon, Jasjeet},
	Journal = {Review of Econonics and Statistics},
	Title = {Genetic Matching for Estimating Causal Effects},
	Year = {2012}}

@book{Wahba:1990,
	Author = {Grace Wahba},
	Publisher = {Society for Industrial and Applied Mathematics},
	Title = {Spline Models for Observational Data},
	Year = {1990}}

@book{Gyorfi2002,
	Author = {Laszlo Gyorfi and Michael Koholor and Adam Krzyzak and Harro Walk},
	Publisher = {New York: Springer},
	Title = {A Distribution-Free Theory of Nonparametric Regression},
	Year = {2002}}

@article{Stone1997,
	Author = {Charles J. Stone and Mark H. Hansen and Charles Kooperberg and Young K. Truong},
	Journal = {The Annals of Statistics},
	Number = {4},
	Pages = {1371--1470},
	Title = {Polynomial Splines and Their Tensor Products in Extended Linear Modeling},
	Volume = {25},
	Year = {1997}}

@article{Hill2011,
	Author = {Jennifer Hill and Christopher Weiss and Fuhua Zhai},
	Journal = {Multivariate Behavioral Research},
	Number = {3},
	Pages = {477--513},
	Title = {Challenges With Propensity Score Strategies in a High-Dimensional Setting and a Potential Alternative},
	Volume = {46},
	Year = {2011}}

@Article{Hainmueller13,
  author  = {Hainmueller, Jens and Hazlett, Chad},
  title   = {Kernel Regularized Least Squares: Reducing Misspecification Bias with a Flexible and Interpretable Machine Learning Approach},
  journal = {Political Analysis},
  year    = {2013},
  volume  = {22},
  number  = {2},
  pages   = {143--168},
}

@article{ratkovic2010finding,
  title={Finding jumps in otherwise smooth curves: Identifying critical events in political processes},
  author={Ratkovic, Marc T and Eng, Kevin H},
  journal={Political Analysis},
  volume={18},
  number={1},
  pages={57--77},
  year={2010},
  publisher={Cambridge University Press}
}

@article{carter2010back,
  title={Back to the future: Modeling time dependence in binary data},
  author={Carter, David B and Signorino, Curtis S},
  journal={Political Analysis},
  volume={18},
  number={3},
  pages={271--292},
  year={2010},
  publisher={Cambridge University Press}
}

@article{Leeb2008,
	Author = {Leeb, Hannes and Benedikt Potscher},
	Journal = {Journal of Econometrics},
	Pages = {201-211},
	Title = {Sparse Estimators and the Oracle Property, or the Return of Hodges Estimator},
	Volume = {142},
	Year = {2008}}

@book{Buhlmann2013,
	Address = {Berlin},
	Author = {Peter Buhlmann and Sara van de Geer},
	Publisher = {Springer},
	Title = {Statistics for High-Dimensional Data},
	Year = {2013}}

@article{Fan:2008,
	Author = {Fan, Jianqing and Jinchi Lv},
	Journal = {Journal of the Royal Statistical Society: Series B},
	Pages = {849--911},
	Title = {Sure independence screening for ultrahigh dimensional feature space.},
	Volume = {70},
	Year = {2008}}

@article{cps2010horseshoe,
	Author = {Carvalho, C and Polson, N and Scott, J},
	Journal = {Biometrika},
	Pages = {465--480},
	Title = {The Horseshoe Estimator for Sparse Signals},
	Volume = {97},
	Year = {2010}}

@article{greenhern,
	Author = {Green, Donald P. and Kern, Holger L.},
	Journal = {Public Opinion Quarterly},
	Pages = {491--511},
	Title = {Modeling heterogeneous treatment effects in survey experiments with Bayesian Additive Regression Trees},
	Volume = {76},
	Year = {2012}}

@article{RatkovicTingley17,
	Author = {Marc Ratkovic and Dustin Tingley},
	Journal = {Political Analysis},
	Number = {25},
	Pages = {1--40},
	Title = {Sparse Estimation and Uncertainty with Application to Subgroup Analysis},
	Volume = {1},
	Year = {2017}}

@article{polson2014bayesian,
	Author = {Polson, Nicholas G and Scott, James G and Windle, Jesse},
	Journal = {Journal of the Royal Statistical Society: Series B (Statistical Methodology)},
	Number = {4},
	Pages = {713--733},
	Publisher = {Wiley Online Library},
	Title = {The bayesian bridge},
	Volume = {76},
	Year = {2014}}

@article{polleysuperlearner,
	Author = {Polley, Eric and van der Laan, Mark},
	Journal = {URL http://CRAN. R-project. org/package= SuperLearner. R package version},
	Pages = {2--0},
	Title = {SuperLearner: super learner prediction, 2012}}

@article{kropko_harden_2020, 
title={Beyond the Hazard Ratio: Generating Expected Durations from the Cox Proportional Hazards Model},
volume={50},
DOI={10.1017/S000712341700045X},
number={1}, journal={British Journal of Political Science}, publisher={Cambridge University Press},
author={Kropko, Jonathan and Harden, Jeffrey J.}, year={2020},
pages={303-320}}

@article{andersen2009nonparametric,
  title={Nonparametric methods for modeling nonlinearity in regression analysis},
  author={Andersen, Robert},
  journal={Annual Review of Sociology},
  volume={35},
  pages={67--85},
  year={2009},
  publisher={Annual Reviews}
}

@article{imai2010general,
	Author = {Imai, Kosuke and Keele, Luke and Tingley, Dustin},
	Journal = {Psychological methods},
	Number = {4},
	Pages = {309},
	Publisher = {American Psychological Association},
	Title = {A general approach to causal mediation analysis.},
	Volume = {15},
	Year = {2010}}

@article{friedman1991multivariate,
	Author = {Friedman, Jerome H},
	Journal = {The Annals of Statistics},
	Pages = {1--67},
	Publisher = {JSTOR},
	Title = {Multivariate adaptive regression splines},
	Year = {1991}}

@article{acharya2015explaining,
	Author = {Acharya, Avidit and Blackwell, Matthew and Sen, Maya},
	Journal = {American Political Science Review},
	Number = {3},
	Title = {Explaining Causal Findings Without Bias: Detecting and Assessing Direct Effects},
	Volume = {110},
	Year = {2016}}

@article{chipman2010bart,
	Author = {Chipman, Hugh A and George, Edward I and McCulloch, Robert E},
	Journal = {The Annals of Applied Statistics},
	Pages = {266--298},
	Publisher = {JSTOR},
	Title = {BART: Bayesian additive regression trees},
	Year = {2010}}

@Article{Sekhon2009,
  author   = {Sekhon, Jasjeet S.},
  title    = {Opiates for the Matches: Matching Methods for Causal Inference},
  journal  = {Annual Review of Political Science},
  year     = {2009},
  volume   = {12},
  number   = {1},
  pages    = {487--508}
}

@book{morgan2014counterfactuals,
	Author = {Morgan, Stephen L and Winship, Christopher},
	Publisher = {Cambridge University Press},
	Title = {Counterfactuals and causal inference},
	Year = {2014}}

@article{AtheyImbens16,
	Author = {Susan Athey and Guido Imbens},
	Journal = {Proceedings of the National Academy of Sciences of the United States of America},
	Number = {27},
	Pages = {7353-7360},
	Title = {Recursive partitioning for heterogeneous causal effects},
	Volume = {113},
	Year = {2016}}

@article{beck1998beyond,
  title={Beyond linearity by default: Generalized additive models},
  author={Beck, Nathaniel and Jackman, Simon},
  journal={American Journal of Political Science},
  pages={596--627},
  year={1998},
  publisher={JSTOR}
}

@Article{aronow2016does,
  author  = {Aronow, Peter and Samii, Cyrus},
  title   = {Does Regression Produce Representative Estimates of Causal Effects?},
  journal = {American Journal of Political Science},
  year    = {2016},
  volume  = {60},
  number  = {1},
  pages   = {250--267}
}

@Book{aronowfoundations,
  title     = {Agnostic Statistics},
  publisher = {Cambridge University Press},
  year      = {2018},
  author    = {Aronow, Peter and Miller, Benjamin},
}

@book{keele2008semiparametric,
  title={Semiparametric regression for the social sciences},
  author={Keele, Luke John},
  year={2008},
  publisher={John Wiley \& Sons}
}

@Book{vand:1998,
  title     = {Asymptotic Statistics},
  publisher = {Cambridge University Press},
  year      = {1998},
  author    = {van der Vaart, Aad},
  volume    = {3},
  series    = {Cambridge Series in Statistical and Probabilistic Mathematics},
}

@Article{chern:chet:2018,
  author  = {Chernozhukov, Victor and Chetverikov, Denis and  Demirer, Mertand Duflo, Esther and  Hansen, Christian and Newey, Whitney and Robins, James},
  title   = {Double/Debiased Machine Learning for Treatment and Structural Parameters},
  journal = {The Econometrics Journal},
  year    = {2018},
}

@Article{wage:athe:2017,
  author  = {Wager, Stefan and Athey, Susan},
  title   = {Estimation and Inference of Heterogeneous Treatment Effects using Random Forests},
  journal = {Journal of the American Statistical Association},
  year    = {2017},
}

@book{imbens2015causal,
  title={Causal inference in statistics, social, and biomedical sciences},
  author={Imbens, Guido W and Rubin, Donald B},
  year={2015},
  publisher={Cambridge University Press}
}

@article{montgomery2018tree,
  title={Tree-Based Models for Political Science Data},
  author={Montgomery, Jacob M and Olivella, Santiago},
  journal={American Journal of Political Science},
  volume={62},
  number={3},
  pages={729--744},
  year={2018},
  publisher={Wiley Online Library}
}

@Article{kenn:lorch:inpress,
  author        = {Edward H. Kennedy and Scott A. Lorch and Dylan S. Small},
  title         = {Robust causal inference with continuous instruments using the local instrumental variable curve},
  journal       = {Journal of the Royal Statistical Society: Series B},
  year          = {In Press},
}

@Article{fong:hazl:2018,
  author        = {Christian Fong and Chad Hazlett and Kosuke Imai},
  title         = {Covariate balancing propensity score for a continuous treatment: Application to the efficacy of political advertisements},
  journal       = {The Annals of Applied Statistics},
  year          = {2018},
  volume        = {12},
  number        = {1},
  pages         = {156-177},
}

@Article{athe:tibs:2019,
  author  = {Susan Athey and Julie Tibshirani and Stefan Wager},
  title   = {Generalized Random Forests.},
  journal = {Annals of Statistics},
  year    = {2019},
  note    = {Forthcoming},
}

@article{breiman2001random,
  title={Random forests},
  author={Breiman, Leo},
  journal={Machine learning},
  volume={45},
  number={1},
  pages={5--32},
  year={2001},
  publisher={Springer}
}

@Article{bech:hain:2011,
  author  = {Michael M. Bechtel and Jens Hainmueller},
  title   = {How Lasting Is Voter Gratitude? An Analysis of the Short- and Long-Term Electoral Returns to Beneficial Policy},
  journal = {American Journal of Political Science},
  year    = {2011},
  volume  = {55},
  number  = {4},
  pages   = {851--867},
}

@Article{leicandes,
  author  = {Lihua Lei and Emmanuel J. Candes},
  title   = {Conformal Inference
of Counterfactuals and Individual Treatment Effects},
  journal = {working paper},
  year    = {2020},
  volume  = {},
  number  = {},
  pages   = {},
}

\clearpage
\spacingset{1.5}

\Large \textbf{Online Appendix for ``Estimation and Inference on Nonlinear and Heterogeneous Effects''}\\\bigskip
\normalsize

\appendix
 
The online appendix for this article contains several sections. Appendix~\ref{app:basisappendix} gives an introduction to using basis functions for regression modelling. Appendix~\ref{app:functionclassappendix} discusses the minimal functional form assumptions made for our model to return consistent estimates. Appendix~\ref{app:pointwise} discusses pointwise bands and contrasts them to the uniform bands we use in the paper for uncertainty estimation.   Appendix~\ref{app:algoverview} gives a detailed exposition of the MDEI algorithm in order to compliment Section~\ref{sec:estimation}. Appendix~\ref{app:sparseappendix} discusses the specific sparse regression model employed in our estimation steps.  Appendix~\ref{app:variance} derives  for our variance estimation.    Appendix~\ref{app:performsims} gives extensive performance simulation evidence leveraging a variety of different data generating processes and comparing MDEI to other relevant methodologies. Finally, Appendix~\ref{app:bincat} discusses the case of a binary or categorical treatment variables rather than the continuous treatment variable context. 

\newpage

\newpage
\section{Introduction to basis functions}
\label{app:basisappendix}

\begin{figure}[t!]
%\vspace{-.9in}
\begin{center}\spacingset{1}
\includegraphics[width=.7\textwidth]{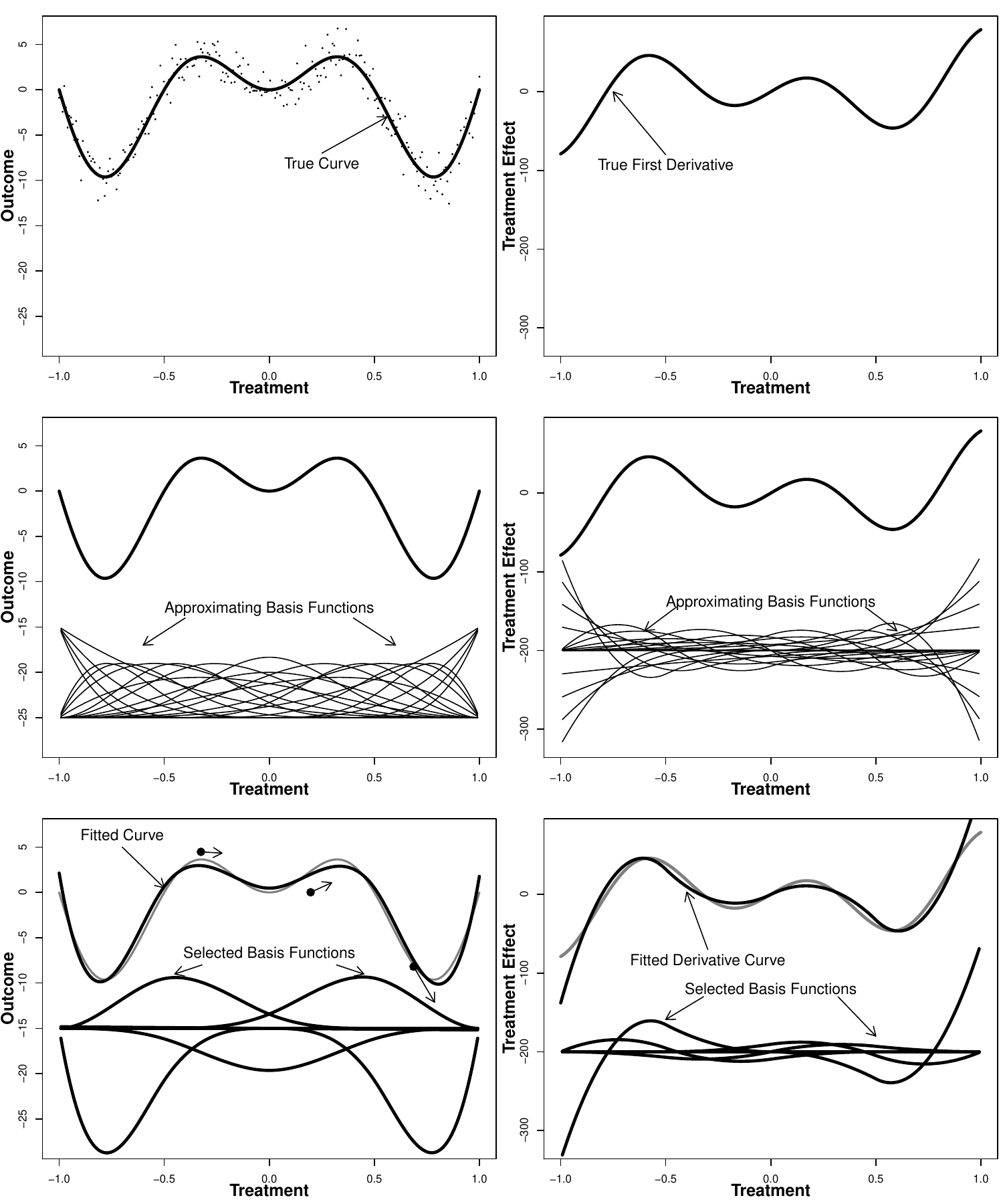}
%BasisExplanation.pdf
\vspace{-.25in}
\end{center}\spacingset{1}
\caption{{\bf Combining Basis Functions to Model Nonlinearities.}This figure illustrates how MDEI uses basis functions to approximate an outcome.  The left column contains results for the outcome $\theta$ and the right for the treatment effect $\tau$.  For simplicity, we assume both are solely a function of the treatment.  The top row shows the true curves to be estimated, with the data in the left.  The middle middle row illustrates the full set of basis functions that we use to approximate the true curve and its first partial derivative.  The bottom row shows the basis functions that were selected, such that when added up they produce the fitted outcome (left) and derivative (right).   A set of random points are selected with the estimated derivative at each point in the left figure.  Rather than assuming a functional form, MDEI uses sets of basis functions that can be combined and interacted to provide an accurate local prediction.
} \label{fig:basis}
\end{figure}

We employ a nonparametric regression model  that models the outcome as an additive sum of a large number functions of the covariates, called {\it basis functions}.   Each basis is a nonlinear transformation of a covariate, allowing us to model a more flexible set of functions.   We illustrate this approach in Figure~\ref{fig:basis}.  The top row shows an example of a nonparametric curve with the data (left) and its first derivative (right).  For simplicity, we make the curves only a function of the treatment.\footnote{For completeness, $T$ is uniform on $[-1,1]$ with $n=250$ and the outcome is $\theta(T)=4\pi T \sin(2\pi T)$ and $\tau(T)=4\pi\sin(2\pi T)+8\pi^2T\cos(2\pi T)$ with normal, mean-zero noise with variance $2$.}

The middle row shows the true systematic component for the outcome and treatment effect and, below it, a set of basis functions that we use to approximate the curve. MDEI uses 23 for each covariate: the linear covariate and then $B$-splines of degree 3, 4, 5, and 6.  These are illustrated below the true curve on the left. Each is differentiable, so their derivatives are shown on the right. These bases are then interacted with each other to approximate ever more complex functions, while we use a variable selection method to select an approximating subset. 

Less important than the particular basis functions is that they are able to approximate a broad set of functions. We illustrate how these basis functions can add up and accurately recover a complex function in the third row. The left shows the estimated conditional mean; the right, the estimated partial derivative.  A few points are selected as well, with their estimated derivative.  We see that these basis functions can recover the relevant trends in the data generating process.
 
\newpage

\section{Function Classes}
\label{app:functionclassappendix}

In this appendix we lay out additional details about the minimal assumptions we make about the sets of function our approach can handle (see Section~\ref{sec:estimation}).

\subsection{Moving beyond parametric functions}
\label{sec:beyondpara}

Extending past the simple linear regression, and more complicated models like the partially linear model, requires several analytic tools.  These tools are necessary to characterize what, exactly, we mean by a complex model; what it means for the estimates of a complex model to be sufficiently ``close'' to the truth to allow for inference; and what sorts of inferential claims we can make about these curves. Each is crucial in moving beyond the linear model.  

For clarity, we illustrate using the function $f$ from our model, which adjusts for covariates in the outcome model.  First, we distinguish between a ``parametric'' and ``nonparametric'' model.  While there is no agreed-upon definition \citep[see, e.g.][p. 1]{wass:06}, the distinction relies on the nature of the underlying assumptions.  Assuming $K$ elements of $X$, say linear terms and any pre-specified interactions or higher order terms, a {\it parametric} model is one where the model is specified in advance, such as
\begin{align*}
     f(X)= X^\top \gamma
\end{align*} 
with $K$ elements in $\gamma$.  We are assuming, then, that $f$ lives in the space of functions linear in the elements of $X$ denoted $X_k$,
\begin{align*}
    \{f:f(X)=\sum_{k=1}^K \phi_k(X) c_k; \; \phi_k(X)=X_k \}
\end{align*}
The {\it basis functions} of a space, which we denote $\{\phi_k\}_{k=1}^K$ , are a set of functions which can be combined to represent any function in the space.  In the parametric setting, the basis functions  $\{\phi_k\}_{k=1}^K$ are simply individual covariates $\{X_k\}_{k=1}^K$. In the nonparametric case discussed below, the basis functions can be more complicated but still combine together to represent some target function like $f(X)$, or, more generically, a conditional mean or even density. As discussed below, different types of approaches to constructing basis functions and estimating their parameters will be used.\footnote{For example, cubic smoothing splines are often used to model time \citep[e.g.,][]{ratkovic2010finding,carter2010back} and other continuous covariates \citep{keele2008semiparametric}. Appendix~\ref{app:basisappendix} provides a comprehensive introduction to regression modelling using  basis functions in the present context.}

The crucial characteristic of a {\it parametric model}  is that the number of parameters in the model (the $K$ parameters  $c_k$) are fixed and finite, and we can rely on asymptotics fixed in $K$ with the sample size $n$ growing.  We adopt the intuition that a {\it nonparametric model} is one where we do not assume the model in advance, and instead the model specification is learned from the data.  There are several ways to consider this in a regression setting.  We may do so by allowing the number of basis functions to grow large, and possibly infinite, in order to accommodate a wide variety of nonlinear, interactive functions in $X$. 

The basis function approach itself subsumes the linear model, meaning that if the model is linear, we recover it, but we also allow include a much larger set of covariates in the regression to pick up unanticipated nonlinearities and interactions.  The cost of this added flexibility is that our asymptotic analysis grows trickier.  Since the number of basis functions, $K_n$, may be larger than the sample size $n$, or even infinite, we can no longer rely on parametric asymptotic arguments. In order to conduct inference, we must characterize our model space precisely and guarantee that we can still recover consistent estimates of the functions within it.

\subsection{Functions for the treatment versus covariates}
\label{sec:treatcovarfunctions}

The \textit{parametric} model will successfully adjust for confounders under the condition that these confounders enter the model linearly and additively.  In other words, in the parametric model, our inference requires the conditional mean be in the function space 
\begin{align*}
\{f: f(X)=X^\top \gamma \}.
\end{align*} We may want to consider the space  
\begin{align*}
\{f: f(X)=\sum_{k=1}^{K_n} \phi_k(X)c_k\},
\end{align*} 
which is similar to the linear space, except we allow the number of bases $K_n$ to grow in $n$ and become potentially infinite, in the limit.

While certainly more flexible than the linear space, we cannot use a finite dataset to estimate an infinite number of parameters \citep{Gyorfi2002}.  We can, though, retain a growing or infinite number of basis functions, a nonparametric space, if we constrain the function space in some manner.  Perhaps the simplest example is the ``sparsity assumption'' that only some finite subset of the parameters $\{c_k\}_{k=1}^\infty$ are not zero, formulating the problem as one where the true model is parametric but we just do not know which basis functions constitute the true model \citep{Buhlmann2013,Belloni14}.  

We implement methods that do not require the sparsity assumption, but constrain the function space so we can still fit models much more complex than a linear model while also recovering a consistent estimate.\footnote{Any consistent regression based method will work within our framework.  A regression framework is necessary, since taking a derivative is straightforward.  We utilize a sparse regression model and give conditions for its consistency below.} Below we consider two different classes of functions.  We will use the first to model our treatment effect, and it will consist of smooth, differentiable functions of the treatment and covariates interacted together.  We will fit this component using a high-dimensional regression model.  The second class we consider is, roughly, functions that we can approximate well using a random forest. We will use this class to model any confounding or bias introduced from the covariates.\footnote{We do not use the same function classes for each since recovering the partial effect with respect to the treatment requires restricting attention to differentiable functions.  We use the random forest for the covariates due to the method's speed and well-established accuracy.} 

\subsubsection{Modeling the Conditional Mean and Partial Derivative}

We first consider modeling $\theta(T,X)$, the part of the outcome explained by the treatment variable. In order to estimate the parameters, we constrain the full function space such that the functions vary, but not too wildly as to render estimation and inference impossible. We do so in three steps.  First, we require $\phi_j(T,X)$ to be bounded.  This allows us to guarantee that no one basis function goes off to infinity, which would leave inference untenable.  Second, since we are interested in modeling ceteris paribus shifts in the treatment on the outcome, we require the basis functions to have a bounded partial derivative in the treatment.  Third, we require the function to be ``simple enough'' that we can recover it from the data.  We do so by requiring the sum of the absolute values of the parameters $\{c_j\}_{j=1}^\infty$ to be finite.  This gives our space for $\theta$ as 
\begin{align*}
     \Theta= \{\theta: \theta(T,X)=\sum_{j=1}^{\infty} \phi_j(T,X)c_j; \phi_j(T,X) \textrm{ and } \frac{\partial}{\partial T} \phi_j(T,X) \textrm{ bounded};\; \sum_{j=1}^{\infty} |c_j|< \infty\}.
\end{align*}
Taking the partial derivative, we can get the space containing $\tau$.\footnote{
\begin{align*}
     \mathpzc T= \{\tau: \tau(T,X)=\sum_{j=1}^{\infty} \frac{\partial}{\partial T}\phi_j(T,X)c_j; \phi_j(T,X) \textrm{ and } \frac{\partial}{\partial T} \phi_j(T,X) \textrm{ bounded};\; \sum_{j=1}^{\infty} |c_j|< \infty\}.
\end{align*}
}

Importantly, this subspace subsumes the linear model.\footnote{This space is a subspace of $L_1(P)$, the space of bounded functions with  finite $L_1$ length of the parameters,  consisting of functions partially differentiable in $T$.  We note that standard results normally require working in $L_2(P)$, which contains $L_1(P)$.  We use $L_1(P)$ since it leads to ``sparser'' estimates and handles the setting with a large number of basis functions better than working in $L_2(P)$. In practice, it is the difference between choosing a LASSO prior and a ridge prior on the basis functions.} For example, if the true model were parametric and linear in basis functions, we would recover the parametric model. 

Before continuing, it is worth a brief mention of what sorts of functions are not in this space.  First are those that are discontinuous functions of smooth covariates, since we only consider smooth bases.  We present just such an example in Section \ref{sec:illustrative}.  We also do not accommodate complex, erratic functions, meaning those such that the sum of the absolute values of the parameters diverges.  For example, if we take $c_j=1/j$, so the parameters have a long heavy tail, our results would not hold.   The closer the model is to sparse, meaning $|c_j|$ decays quickly or even becomes zero, the better we expect our method to perform.

\subsubsection{Modeling the Nuisance Functions  using a Lipschitz Space}
We turn next to modeling  how the covariates affect the outcome and treatment, as represented by the functions $f,g$ respectively. Here, we simply assume that these functions can be well-estimated using a random forest, which places them in a {\it Lipschitz space}.\footnote{This is the space of functions where the slope of any secant line between any two points is bounded by some constant, say $C$. To formalize, this space can be characterized as 
\begin{align*}
Lipschitz(\alpha)=\{f: |f(X)-f(X^\prime)|\le C |X-X^\prime|^\alpha\textrm{ for some }C<\infty\},
\end{align*}
This is the most general space we use, where $\textrm{Linear spaces} \subset \Theta \subset \textrm{Lipschitz functions}$.} In essence, by using random forests to partial out the covariates--as part of an estimation and inference procedure--we can be operating in the seminonparametric framework.

This Lipschitz assumption is necessary to allow us to use random forests to adjust for the covariates. It is also more general than the space we use for treatment $\times$ covariate interactions, since Lipschitz functions are continuous but need not be differentiable.  We add more structure to the space where we look for treatment effects, which we operationalize as a derivative in the case of continuous treatment variable.

\section{Pointwise Confidence Bands}
\label{app:pointwise}

A second, and subtly different idea of coverage, is the pointwise band.  This band is asymptotically valid at every point, 
\begin{align}
 \inf_{\substack{ \tau \in \mathpzc T\\f,g \in Lipschitz(1)\\H \in \mathpzc H}}  \lim_{n \rightarrow \infty}\Pr_{\mathcal D_n}(\tau(T^\prime,X^\prime) \in CB_{\tau; \mathcal D_n}(T^\prime,X^\prime) \;\forall\; (T^\prime,X^\prime)  \in \mathcal T \times \mathcal X) ) \ge 1-\alpha
\end{align}
which is the uniform band except with the $\lim$ and the $\inf$ reversed.  This is the form of inference most commonly encountered and easiest to establish; see, e.g., Theorem 5 of \citet{athey2019generalized}.  This confidence band is constructed to be asymptotically valid at every single point.  This means, given a point $(T^\prime, X^\prime)$, we would expect the pointwise confidence band to contain the true value $\tau(T^\prime, X^\prime)$ at this point at least $100\times(1-\alpha)\%$ of the time, over repeated samples.

\subsection{Pointwise versus Uniform Bands}

The uniform band is to be preferred \citep[e.g.,][]{wass:06}. The argument is subtle but worth explaining.  A uniform band ensures some statistical property over the whole of the curve.  An asymptotically uniform band, then, guarantees there exists a sample size $n_0$ such that for all sample sizes $n>n_0$, the band is valid.  The pointwise band offers no such guarantee over the whole of the curve, but just that the band is valid at every point.

For example, a $100\times(1-\alpha)\%$ uniform band is asymptotically valid if it contains the entire true curve across  $100\times(1-\alpha)\%$ of repeated samples, in the limit.  A  $100\times(1-\alpha)\%$  pointwise coverage if, at every single point, it contains the true value at this point over $100\times(1-\alpha)\%$   of repeated samples, in the limit.

These two claims regularly diverge in nonparametric estimation for a subtle reason: not every point along a nonparametric curve will converge at the same rate in sample size.  Recall that in order to identify the model, we need to restrict our attention to a particular space.  The estimate will converge faster in areas where it is closer to our assumed space, and slower in other spaces.  For one example, \citet{Leeb2008} show that if you work under a sparsity assumption--that only a finite number  of the parameters $c_j$ are non-zero--you can recover standard parameteric pointwise confidence intervals on each coefficient that shrink at the rate $n^{-1/2}$.  These intervals are only valid if the model is in-truth sparse, but fall apart otherwise.   If there are parameters that converge to zero at a rate of $n^{-1/4}$, the pointwise confidence interval can be arbitrarily misleading, since it may be missing parts of the true curve by a non-negligible amount that will leave our inference asymptotically invalid.  Driving the distinction is that, while we may be able to make inferential claims about a given point on a curve, this is not the same as making such a claim along the curve.

\section{Overview of Algorithm}
\label{app:algoverview}

\subsection{Overview}

In estimating and conducting inference on $\tau$, we have three core goals: to adjust for confounders nonparametrically, to allow the instantaneous effect of the treatment to be a nonparametric function of the treatment and interactions with covariates, and to recover uncertainty estimates that allow for inference. Given these challenging goals, we must use a number of steps including: a split-sample approach (to avoid over-fitting), partialing out covariates via a random forest (to deal with confounding nonparametrically), a screening step and regression model with nonparametric basis functions (to capture any complicated interactions between the treatment and covariates, as well as to allow the impact of the treatment on its own to be nonlinear), conformal inference methods (to recover a uniform confidence band), and finally repeated cross-fitting (to reduce sensitivity to the particular splits of the data we use at each stage). In what follows we lay out our model of the treatment effect and then work through each of these steps. Appendix~\ref{app:algoverview} provides a formal algorithm diagram.

\subsection{The Implemented Treatment Effect Model} \label{app:implementedmodel}

Recall the model that we are estimating is 
\begin{align*}
    \begin{split}
    Y=\theta(T,X) +f(X) +\epsilon,\;\; \E(\epsilon|T,X)=0\\
    T= g(X)+u,\;\; \E(u|X)=0.
        \end{split}
\end{align*}
and our target is the function $\tau(T,X) = \frac{\partial}{\partial T}\theta(T,X)$.  As discussed above, doing so requires making modeling choices for each component.  We assume that $f, g$ are in a broad class of functions estimated consistently by a random forest, denoted $Lipschitz(1)$.  We place $\theta$ in a class of bounded, differentiable functions--so that we can recover $\tau$--such that we can estimate it consistently through a regression model with nonparametric basis functions.

Before proceeding with our exact implementation procedure we first describe the model of $\tau$. We will use a set of nonparametric bases, denoted $\phi_j$, where each is a nonlinear function of its input. We use a specific set of bases used in smoothing splines, the $B$-spline basis, which are a set of bounded, nonlinear transformations commonly used to model complex data.\footnote{Appendix~\ref{app:basisappendix} provides a more comprehensive introduction to regression modelling using basis functions.} Specifically, we use degree 3, 4, 5, and 6 degree B-spline bases, for a total of 23 bases. We draw on smoothing splines for a very specific reason: we want to allow for the possibility that $\tau(T,X)$ is non-linear in $T$. For example, the effect of the treatment could be quadratic but given our seminonparametric approach we will not pre-specify that possibility in advance and simply let the data determine the relationship.    

We then interact this set of  nonparametric bases, denoted $\{\phi_j\}_{j=1}^{23}$, where each basis beyond the first  is a nonlinear function of its input.  We interact these in order to accommodate functions that are nonlinear, possibly even nonmonotonic, and interactive in the covariates.\begin{footnote}{If $K=5$ there will be over $175,000$ potential basis functions and $K=10$ generates over $6,900,00$.  This explosion generates one of our greatest estimation challenges. These bases constitute a larger space than those considered by earlier multivariate nonparametric regression models that inspire our work; see \citet{Stone1997,friedman1991multivariate} for seminal work and \citet{Ravikumar09} for recent work.}\end{footnote} Because we will be partialing out the $X$'s, we model $\theta$ in terms of $\widetilde T=T-\widehat \E(T|X)$.  

Then, for elements of $X=[X_1, X_2, \ldots, X_k, \ldots, X_K]$, we model all $treatment \times covariate \times covariate$ interactions, giving us the model for $\theta$:
\begin{align}
\begin{split}
\theta(\widetilde T,X)=& \widetilde T \gamma_0+  \\
&\sum_{k=1}^K  \sum_{{k^\prime}=k}^K  \sum_{j=1}^{24}\sum_{j^\prime=1}^{24}\sum_{j^{\prime\prime}=j^\prime}^{24} \phi_j(\widetilde T) \phi_{j^\prime}(X_k) \phi_{j^{\prime\prime}}(X_{k^\prime}) c_{k k^\prime jj^\prime j^{\prime\prime} }.
\end{split}
\end{align}
Here, the terms $c_{\cdot}$ are  simply parameters in a regression model attached to each of the basis functions.

We estimate $\theta$ directly, but are able to infer $\tau$ by taking the partial derivative with respect to $\widetilde T$. This gives us a model for $\tau$ as
\begin{align}
\tau(T,X)=  \gamma_0+ \sum_{k=1}^K  \sum_{{k^\prime}=k}^K  \sum_{j=1}^{24}\sum_{j^\prime=1}^{15}\sum_{j^{\prime\prime}=j^\prime}^{24} \frac{\partial}{\partial \widetilde T} \phi_j(\widetilde T) \phi_{j^\prime}(X_k) \phi_{j^{\prime\prime}}(X_{k^\prime}) c_{k k^\prime jj^\prime j^{\prime\prime} }.
\end{align}

\subsection{Full Sample Estimation Stage}

Having stated our model of $\tau$ we are now ready to work through our estimation procedure. The first stage uses the full sample of data. We want to identify a large number of nonlinear transformations of $X$ and $T$ that we can use to model $\theta$ and, consequently, $\tau$.  Our goal in this preliminary step is to simply generate a large set of bases that might help model the true $\theta$ and, hence, $\tau$. We are not committing to a model here, but a model {\it space}, i.e., we are simply trying to recover a set of several hundred basis functions that might be informative for the outcome.  We then use these basis functions in the subsequent estimation stages. We proceed next with each of the full sample steps.

\paragraph{Full Sample Step 1: Partialling out covariates} \label{app:fullsample}

Our first step is to partial out the covariates from the treatment and the outcome. To do this, we calculate the expected value of $Y$ and $T$ given the covariates $X$ and then subtract them from the observed values to produce two sets of residuals. 
\begin{align}
    \widetilde Y =Y - \widehat \E(Y|X);\;\; \widetilde T =T - \widehat\E(T|X)
\end{align}
Throughout, when we partial out the covariates, we use a random forest to model the conditional expectation of $Y$ and $T$ \citep{breiman2001random,montgomery2018tree}.\footnote{We use the implementation from the {\bf R} package {\tt randomForest}.  We chose this implementation due to its speed and ubiquity.} A random forest is particularly helpful in doing this because it is flexible, fast, and does not make strong assumptions about the conditional mean. We will directly model these residuals having partialed out the covariates.

\paragraph{Full Sample Step 2: Iterated SIS Screen}

We next turn to selecting a set of bases we can use to model $\theta, \tau$.

\paragraph{SIS Screen Iteration 1}
The first, screening, step is designed to get us from a potentially vast number of bases to a more manageable number.  As originally presented in \citet{Fan:2008}, the Sure Independence Screen (SIS) selects a subset of bases that have the highest correlation with the outcome. Their method assumed the data are jointly normally distributed, in which case the correlation is an appropriate measure of the relationship between an outcome and basis. We produce a more stable and reliable screening procedure by using a  measure of correlation that is robust to outliers and peculiarities of the full sample.\footnote{To do so, first, we split the data in half, measure the correlation between outcome and basis on each half, and repeat this process five times.  We then only consider bases that have the same sign on the correlation with the outcome across all splits. Second, having dropped bases with unstable relationships, we select those where the median value of the correlation across the splits is as large as possible in magnitude. In this way, we are  able to guard against correlations that are not robust across splits of the data, and guarantee that the median correlation is still sizeable.}  In this process, we select $20\times(1+n^{1/5})$ bases that have the highest robust correlation with $\widetilde Y$.\footnote{
It is important to note that, in our implementation, we are implicitly assuming that the true function $\theta$ can be captured by our one hundred recovered bases.  This is not a theoretical requirement--we could have the number of bases grow in sample size--but instead an implementation decision that we find works well in practice. If we wish to recover a true sparse model, we need a ``beta-min'' assumption on the smallest true value.  We do not require consistency in variable selection, but consistency in prediction, which requires the much weaker assumption that the sum of the $L_1$ norm of the mean parameters is finite; see \citet[][ch. 6]{Buhlmann2013}.}   

\paragraph{Intermediate Regression Step}

A sparse regression (e.g., LASSO) is then fit using the bases returned from our modified SIS screen and the outcome and treatment variables that had been previously residualized by partialling out the covariates (i.e., $\widetilde Y, \widetilde T$). To do so, we take our outcome $\widetilde Y$, and use our Bayesian regression model to regress it on the bases selected in this first stage of the screen (denoted $bases_1$), producing residuals
\begin{align}
    \widetilde Y^{SIS1}=\widetilde Y-\widehat \E^{LASSO}
    (\widetilde Y|bases_1),
\end{align}
which we enter into the next SIS iteration.

\paragraph{SIS Screen Iteration 2}

As noted in the original work of \citet{Fan:2008}, iterating the screening process is effective can improve our estimation accuracy.  
We use the values from the previous step, $\widetilde Y^{SIS1}$, to select a new set of one hundred bases. To do this, we repeat the process as above, except with $\widetilde Y$ replaced by $\widetilde Y^{SIS1}$. This additional set of $20\times(1+n^{1/5})$  gives us one more chance to incorporate relevant bases that may have been missed in the first screen, producing $40\times(1+n^{1/5})$ bases.

\paragraph{Summary} Our iterative SIS process produces $40\times(1+n^{1/5})$  bases for $\theta(T,X)$ that have the property that they have a stable correlation with the outcome. We will then use these bases below for the actual estimation and inference steps.\footnote{Since this is our final iteration of the screen, we do not need to regress the residualized outcome $\widetilde Y^{SIS1}$ onto the bases selected at this stage.}\footnote{We pass through to the auxiliary and estimation sample stages the linear interactions between the treatment variable and covariates. Below when we estimate our quantities of interest using our high dimensional regression model, we still allow these terms to be zeroed out. That is, we always leave them under consideration rather than screening at this full sample stage.} 

\subsubsection{Full Sample Summary}

From this full sample portion, we recover a set of candidate basis functions that we will use to model $\theta, \tau$ below.  Our aim was not to select, but to screen--meaning getting the number of bases from hundreds thousands to low hundreds. We then take these low-hundreds set and further select a useful subset (auxiliary sample below) and use these to construct a confidence interval (estimation sample below).

\subsection{Split-sample Estimation Stage} 

In order to avoid any accidental bias due to the over-fitting of our our machine learning method, we implement a {\it split-sample} strategy. To do so, we first split the sample into two equally sized splits: an {\it auxiliary sample} and an {\it estimation sample}.  This split-sample approach serves as a crucial guard against data dredging, since we use the auxiliary  sample of the data to learn models of the covariates, but the estimation sample to conduct inference \citep[see, e.g.,][]{vand:1998, AtheyImbens16, chern:chet:2018}.  This provides a guard such that any incidental over-fitting on the auxiliary split will not carry through to our calculations on the estimation split.  

In particular, we split the sample in two, and use the first half to adjust for background covariates and to learn a plausible specification for $\theta$.  We then turn to the second half to estimate $\widehat \theta$ and, consequently, $\widehat \tau$.  We then use this split of the data to construct a confidence band around $\widehat \tau$.  We estimate the standard errors and select a critical value such that the interval contains $100\times (1-\alpha)\%$ of the observed data.

\subsubsection{Auxiliary Sample}

\paragraph{Auxiliary Sample: Partialing out covariates using a random forest}
 
Using the auxiliary sample, we again partial the covariates out of the treatment and outcome:
\begin{align}
\widetilde Y= Y-\widehat \E^{aux}(Y|X); \widetilde T= T-\widehat \E^{aux}(T|X)
\end{align}
where the $aux$ subscript denotes that the expectation was estimated on the auxiliary sample. We want to partial out the covariates from the outcome and treatment so we can estimate $\theta$ without worrying about the confounding functions $f,g$.  Doing so is also a crucial part of generating narrow confidence bands. Note that since we are using the auxiliary sample to partial out the covariates, the models do not depend on data from the estimation sample, which will allow us to recover the confidence interval.

\paragraph{Auxiliary Sample: Selecting Bases with Sparse Regression}

Given the nonparametric bases that survived the screen, we next use a sparse regression model to select a subset for modeling $\theta$.    A {\it sparse regression} is a regression that returns point estimates of zero for all but a subset of covariates; we use a slight adaptation the method presented in \citet{RatkovicTingley17}.Appendix~\ref{app:sparseappendix} discusses the particulars of the model.\footnote{For a more complete discussion of sparse regression methods, see the discussion in \citet{RatkovicTingley17} and \cite{Buhlmann2013}. } Sparse regression offers the ability to select a subset of relevant predictors and estimate their coefficients at the same time.  Doing so, though, introduces a non-negligible bias--again, learning the model and estimating its coefficients off one data set can introduce a bias that makes inference challenging.

To avoid this problem we use the sparse regression in the auxiliary sample to select a subset of nonparametric bases.  We will take these selected bases from the auxiliary sample and apply them to the estimation sample in order to estimate coefficients and confidence bands.    The important part here is that we separate learning a subset of bases and estimating their coefficients; doing so serves a crucial role in producing valid confidence bands.

\paragraph{Auxiliary Sample: Modeling the Error Variance } \label{para:errorvariance}

One of our central contributions is introducing not just a flexible model of the treatment effect $\tau$ but also confidence bands on its estimate.  We proceed in two steps. First, we generate an upper bound on $\E((\widehat \tau-\tau)^2)$.  A central part is considering the case where $\widehat \tau$ does not converge to $\tau$, due to model misspecification of some type.  We denote the function to which $\widehat \tau$ {\it does} converge as $\widetilde \tau$, and incorporate this into our bounds. Second, we describe a means to estimate it.  This bound, as we describe below, has an important asymptotic property: if the model for $\tau$ is not properly specified, we expect average coverage of our interval to be nominal; if it is properly specified, we expect average coverage to be more than nominal but still tight.

Our strategy relies on generating a bound on our estimation error on $\theta$ and then recovering one for $\tau$.  Denote as $\widetilde \theta$ the limiting value of $\widehat \theta$,
\begin{align}
\widetilde \theta(T,X)=\lim_{n \rightarrow \infty} \widehat \theta(T,X),
\end{align}
i.e., what we would expect $\widehat \theta$ to converge to with an infinite amount of data.  

An important element of this construction is that we are using the conformal band to recover a conservative band on the true value.  This band {\it could} be shortened by making, say, parametric assumptions on the functional form or distributional assumptions on the error term. Instead, we are striving for the narrowest possible bounds that will work regardless of the true model, so as not to outright rely on these assumptions.

Our construction follows two steps; we place the details in Appendix \ref{app:variance}. First, we start with a valid conformal band.  Specifically, given a future realization $Y_i^\prime$, we can generate a bound on $Y_i^\prime-\widehat \theta(T_i,X_i)$, such that an interval centered on the fitted value will contain the future realization with some controlled probability.  Next, we decompose this future observed value into its systematic and random component, $Y_i^\prime=\theta(T_i, X_i) +\epsilon_i^\prime$. Rearranging terms, we are able to bound  the gap between the conditional mean, $\theta(T, X)$, and its estimate $\widehat \theta(T,X)$, as a function of the error variance.  Briefly, we show that if the conformal bound is $c$ at this point, a value of $c+1$ will generate a bound that contains the true value $\theta(T,X)$ with the same probability.  Third, we then take derivatives and generate a similar bound on $\tau(T,X)$, adding one to the critical value.

\subsubsection{Estimation Sample}

From the auxiliary sample, we have gathered: random forest models for how the covariates affect the treatment and outcome; a set of bases selected by the sparse regression for modeling $\theta$; and models for the error variance components. We now take these elements and turn to the estimation sample to construct an estimate.

\paragraph{Estimation Sample: Partialing out Covariates}  We take the random forest model from our auxiliary sample and use it to partial out the covariates in the estimation sample.  Denote as $est$ observations in the estimation sample and $aux$ a model fit on the auxiliary sample, this becomes
\begin{align*}
    \widetilde Y^{est}&=Y^{est}-\widehat \E^{aux}(Y^{est}|X)\\
    \widetilde T^{est}&=T^{est}-\widehat \E^{aux}(T^{est}|X)\\
\end{align*}
    We then use these values throughout.

\paragraph{Estimation Sample: Estimating Coefficients} 

We are now ready to estimate $\widehat \theta$ and, consequently, $\widehat \tau$.  Turning to the estimation sample, we regress $\widetilde Y^{est}$ on the bases evaluated at $(\widetilde T^{est}, X)$.  Importantly, $\widetilde Y^{est}$ and $\widetilde T^{est}$ have had $X$ partialed out using a model fit to the auxiliary data but evaluated on the estimation sample. Similarly, we use the estimation sample to estimate coefficients on the selected bases (we used the auxiliary sample to select these bases).  Regressing $\widetilde Y^{est}$ on the selected bases gives us the coefficients we need to estimate $\widehat \theta, \widehat \tau$. It also gives us estimates of the error variance and residuals that we use to estimate the sampling variance.

\paragraph{Estimation Sample: Estimating the Error Variance}  Estimating the error variance for $\widehat \theta$ and $\widehat \tau$ follows a similar procedure to that for the coefficients. By using the models from the auxiliary sample, we predict the variances on the estimation sample.  

\paragraph{Estimation Sample: A Conformal Method}

We are almost ready to estimate our confidence interval on $\widehat \tau$.  We have reasonable estimates for every component save the critical value $C$, which controls how many standard errors above and below the estimate we extend our confidence band. In order to estimate $C$, we use a simplified version of a data-driven means to select $C$. We adopt the {\it conformal inference} approach (see discussion in Section~\ref{sec:uncertainty} and \citet{lei:sell:rina:tibs:wass:19} for an overview) to generate exact, or nearly exact, confidence intervals that do not rely on normality assumptions. This allows us to recover asymptotically valid confidence bands, yet still have them not rely on distributional assumptions on the error term.

We use a conformal approach to select a critical value such that the symmetric confidence band on $\widehat \theta(T,X)$ contains $100\times(1-\alpha)\%$ of the data:
\begin{align}
\widehat C_{1-\alpha/2}=\inf \{C: \frac{1}{n} \sum_{i=1}^n \mathbf 1\left(Y_i \in CI(\widehat \theta(T,X),\widehat E((\widehat \theta-\theta)^2)|T,X), \mathcal D_n,\widehat C, \alpha)\right)\ge 1-\alpha\}
\end{align}

We then carry this critical value  from the outcome to the treatment effect and recover confidence interval
\begin{align}
CB_{\widehat \tau,\mathcal D_n} = \widehat \tau(T,X) \pm \widehat C_{1-\alpha/2} \sqrt{\widehat s^2_\tau(T,X) +\widehat \sigma^2_\tau(T,X)}.
\end{align}

This conformal approach offers us a principled, and data-driven, means to select a critical value on our confidence band. The value is selected to guarantee average coverage, asymptotically, such that we expect the true curve $\tau$ to fall in the band at about $100\times(1-\alpha)\%$ of the observed data.  This provides a useful tool, mathematically and visually, for characterizing uncovered heterogeneities in our data.

\subsection{Repeated Cross-Fitting}\label{sec:repcrossfit}

One concern might be that the procedure is inefficient since we are splitting the data. Further, it might be sensitive to which splits of the data we are using at each stage.  To ameliorate these concerns, we engage in {\it repeated cross-fitting}.  After using the auxiliary sample to select covariates and the estimation sample to estimate them, we switch their roles, cross-fitting the model to these two splits.  We then repeat this process and average over repetitions.

\subsection{Summary}

In sum, we present a method that offers both advantages and  disadvantages.  The advantage is that we are able to recover a nonparametric treatment effect  with point estimates competitive with several cutting-edge methods.  More important, from our point of view, is the uncertainty estimation.  We are able to recover a confidence band that contains the true value over at least $100\times(1-\alpha)\%$ of the data points, asymptotically.    

We are able to achieve this while making minimal assumptions over the functional form of the treatment effect or the error distribution.  This comes at a real cost: our bands are conservative for two separate reasons.  First, we make only mild functional form or distributional assumptions.  These bands can be narrowed without making functional form or distributional assumptions, but in this manuscript we strive for maximal agnosticism.  Second, the bands are constructed from bounds on asymptotically exact (i.e. non-conservative) bands, so we expect slippage through the bounding.  At the same time, any shrinkage on these intervals would have to come through assumption.

\subsection{Algorithm Diagram}

In this section, we outline three algorithms that we use to implement our method. In the first, Algorithm \ref{alg:bases},  we generate a set of bases that model heterogeneity.  The second, Algorithm \ref{alg:interval}, details how we construct our intervals, given fitted values and standard errors at each point. The third, Algorithm  \ref{alg:estimation}, uses the first two  to construct our estimates.

%%Algorithm for bases
\begin{algorithm}[H]
\KwData{Outcome vector $Y$, treatment vector $T$, covariate $N\times (K+1)$ covariate matrix $X$ with intercept in first column
}
\SetKwInput{KwFunction}{Functions}
\KwFunction{Spline functions $\phi_{jm}$ the $j^{th}$ $B$-spline of degree $m$, standardized.

$\rho(a,b)$ the absolute value of the median of the correlation of $a$ and $b$ taken over 5 splits of half the data.}

\KwResult{A set of indices generating nonparametric bases for modeling treatment $\times$ covariate interactions}
Using the full sample, generate $\widetilde Y=Y-\widehat \E(Y|X),\widetilde T=T-\widehat \E(T|X)$ using random forests \;
Standardize $\widetilde Y, \widetilde T$

 \For{ $j$ in 1 to (K+1)}{
  \For{ $j^\prime$ in 1 to (K+1)}{
  \For{ $d$ in \{3,5,7,9\}}{
  \For{ $d^\prime$ in \{3,5,7,9\}}{
    \For{$m$ in 1 to d}{
        \For{$m^\prime$ in 1 to d}{
        Save $\rho(\widetilde Y, \widetilde T\times \phi_{md}(X_j)\times  \phi_{m^\prime d^\prime}(X_{j^\prime})     $
        }}}}}}
    Return indices corresponding with bases with top $20(1+n^{.5})$ values of $\rho$
 \caption{Generating Candidate Bases}
 \label{alg:bases}
\end{algorithm}
\newpage
    
%% Algorithm for CIs
\begin{algorithm}[H]
\KwData{Matrix of fitted values and estimated standard errors at each point\;
False positive rate $\alpha$
}

\KwResult{Fitted values, first derivative,  variance estimates, and uniform confidence interval }

\label{alg:interval}
 \For{i in 1 to numsplits}{
    Find smallest critical value such that symmetric confidence interval contains $100 \times (1-\alpha)\%$ of the data
 }
Return mean critical value
\caption{Generating Critical Value}
\label{alg:cvalpha}
\end{algorithm}

%% Algorithm for Estimation
\begin{algorithm}[H]
\KwData{Outcome vector $Y$, treatment vector $T$, covariate $N\times (K+1)$ covariate matrix $X$ with intercept in first column, indices $\mathcal I$ for interactive splines, split sample $\mathcal S_0, \mathcal S_1$
}

\KwResult{Fitted values, first derivative,  variance estimates, and uniform confidence interval }
Generate bases $\mathcal I$ from Algorithm \ref{alg:bases}\;

\For{1 in 1 to $numsplits$}{
    Generate $\widetilde Y=Y-\widehat \E^{\mathcal S_0}(Y|X),\widetilde T=T-\widehat \E^{\mathcal S_0}(T|X)$ using random forests but a model fit using $\mathcal S_0$  \;
    Construct heterogeneity bases from $\mathcal I$\;
    Using data from $\mathcal S_0$ select a subset of the bases using a sparse regression;
    Using data from $\mathcal S_0$ model conditional variance by regressing the squared errors on covariates using random forests;
    Regress $\widetilde Y_i$ on selected bases on split $\mathcal S_1$\;
    Evaluate point estimate, first derivative, and variance of fitted and first derivative at each point in $\mathcal S_1$ using selected bases \;
    Generate predicted values of error variance at each point\;
    Switch roles of $\mathcal S_0, \mathcal S_1$ and cross-fit.
    Save fitted values,
    first derivative estimates, variance estimates.
    }
    Calculate uniform confidence intervals using confidence interval using Algorithm \ref{alg:cvalpha}.
 \caption{Estimation Algorithm}
 \label{alg:estimation}
\end{algorithm}

\section{Sparse Regression Model}
\label{app:sparseappendix}

Even after screening, we still have hundreds of nonparametric bases.  Regressions of this magnitude, though, can be estimated reliably using existing high-dimensional regression methods.  We implement the high-dimensional regression described by  \citet{RatkovicTingley17}.  This work was focused on variable selection, estimating a subset of bases that are likely non-zero.  Our problem is subtly different: we want the best predictive model.  

High dimensional regression requires a tuning parameter, $\lambda$, that controls the level of shrinkage.   We implement an adaptation of the Bayesian sparse regression, from \citet{RatkovicTingley17}.  For completeness, we present the full model hierarchy,

\begin{align}
Y_i |X_i, \beta, Z_i, b\sigma^2 &\sim \mathcal N(X_i^\top \beta, \sigma^2)\\
\beta_k| \lambda, w_k, \sigma &\sim DE\left(\lambda w_k/\sigma\right)\\
\lambda^2|N, K& \sim \Gamma\left(\alpha, \rho\right) \\
w_k|\gamma &\sim \textrm{generalizedGamma}(1,1,\gamma)\\
\gamma &\sim \exp(1) \\
%b|\sigma^2_1, \ldots, \sigma^2_G &\sim \mathcal N(0_{|b|}, V_b)\\
%\sigma^2_g &\sim InverseGamma(0,1)
\end{align}

Taking $\alpha = \sqrt N K$ gives the model of \citet{RatkovicTingley17}. In that model, the goal was consistent model recovery, meaning zeroing out irrelevant covaraites and maintaining relevant ones.  This required selecting $\alpha$ so that the model would successfully distinguish between the two types. We instead adjust the prior on $\lambda$ to $\alpha=n \log p$, which will optimize prediction, rather than model recovery.  They differ in that, for prediction, we are more willing to tolerate irrelevant variables so long as they help improve prediction.  Setting $\lambda$ to  the  ``oracle rate'' $\lambda = O\left(\sqrt{n \log p}\right)$  will minimize the model's predictive error \citep[see, e.g.][ch. 6]{Buhlmann2013}.  In this manuscript, we are analyzing predicted values, not selected variables.  To this end, we select our tuning parameter for optimal  prediction.

\section{Variance Derivation}\label{app:variance}

\subsection{Deriving the Conformal Bound}

We assume we have a valid conformal bound.  Then, for some future value $Y_i^\prime$ with covariates $X_i$, variance at this point $\widehat \sigma_{\widehat\theta}(X_i)$, and critical value $\widehat C_{1-\alpha/2}$, we get
\begin{align}
    \Pr_X(| Y_i^\prime-\widehat \theta(X_i)| \le \widehat C_{1-\alpha/2} \widehat \sigma_{\widehat\theta}(X_i)) \ge 1-\alpha.
\end{align}
Now, we bound the inequality inside the probability from the left using
%$\frac{1}{2}(a+b)^2 -b^2 \leq a^2$
$|a+b|-|b| \leq |a|$
where $a+b=\widehat \theta(X_i)-\theta_i$, $a=Y_i^\prime-\widehat \theta(X_i)$ $b=\theta_i-Y_i^\prime$.
So, with a conformal band, we can ensure the following event occurs with probability at least $1-\alpha$
\begin{align}
   \widehat C_{1-\alpha/2} \widehat \sigma_{\widehat\theta}(T_i,X_i) &\ge | Y_i^\prime-\widehat \theta(T_i,X_i)^\prime|\\
   &\ge  |\widehat \theta(T_i,X_i)-\theta_i| -|\theta_i-Y_i^\prime|
\end{align}
and rearranging, then bounding the term on the right gives
\begin{align}
    |\widehat \theta(T_i,X_i)-\theta_i)|&\le \widehat C_{1-\alpha/2} \widehat \sigma(T_i,X_i) +|\theta_i-Y_i^\prime|\\
    &\le (\widehat C_{1-\alpha/2} +1)\widehat \sigma_{\widehat\theta}(T_i,X_i)
\end{align}
since $\widehat \sigma(X_i) \ge |\theta_i-Y_i^\prime|$, in expectation. 

Thus, we should expect the confidence band 
\begin{align}
    CB_{\widehat \theta,\mathcal D_n} = \widehat \theta(T_i,X_i) \pm (\widehat C_{1-\alpha/2} +1)\widehat \sigma_{\widehat\theta}(T_i,X_i)
\end{align}
to have at least $100\times(1-\alpha)\%$ average coverage of the systematic component $\theta(T,X)$.  We replace the conformal critical value $\widehat C_{1-\alpha/2}$ with $\widehat C_{1-\alpha/2}+1$, which has to be widened to better include $\theta$, rather than a future predictive value.

Since our bounds are not exact, we expect it to be conservative for $\theta(T_i,X_i)$. The predictive bound is exact, asymptotically, but our bound on the true systematic component comes from bounding this conformal band.  Therefore, we expect the coverage of the $100\times(1-\alpha)\%$ band to be greater than $100\times(1-\alpha)\%$, but this is the cost we had to incur in moving from bounding the predictive value to the systematic component.

We then use this critical value to construct a band around $\widehat \tau$ as 
\begin{align}
    CB_{\widehat \tau,\mathcal D_n} = \widehat \tau(T_i,X_i) \pm (\widehat C_{1-\alpha/2} +1)\widehat \sigma_{\widehat\tau}(T_i,X_i)
\end{align}

We estimate the variance using the law of total variance
\begin{align}
   \underbrace{\widehat \sigma_{\widehat\theta}^2(T_i,X_i)}_{\textrm{Total Variance}}= \underbrace{\widehat s_{\widehat\theta}^2(T_i,X_i)}_{\textrm{Sampling Variance}}+\underbrace{ \widehat \sigma_{\theta}^2(T_i,X_i)}_{\textrm{Error Variance}}
\end{align}
We calculate the sampling variance as the variance in the fitted values over repeated cross-fits.  We then estimate the error variance using a random forest on the squared residuals, and these estimates are also averaged over split-samples.

We then construct our error on $\widehat \tau$ using the same formula.  The sampling variance can, again, be calculated from the cross-fit sample variance over the estimates.  For the second variance term, we estimate the variance of $Y$ attributable to $T$, but not $X$, which we estimate as
\begin{align}
\widehat \sigma^2_{\widehat \tau(T,X)} = \widehat \Var^{aux}(Y|X)-\widehat \Var^{aux}(Y|T,X).
\end{align}
where the estimates are constructed using random forests on the auxiliary sample.  In the limit, this estimate should be nonnegative; in practice, we instead take its absolute value.

\section{Performance Simulations}
\label{app:performsims}

\subsection{Data Generating Processes}

We next present simulation evidence illustrating MDEI's utility in estimating treatment effect.  We include four sets of simulations presented in increasing complexity, a linear model, a low-dimensional interactive model, a high-dimensional interactive model, and a model with a nonlinearity, respectively:

In each setting, we generate five covariates $X_{i1}, \ldots, X_{i5}$ from a standard multivariate normal with correlation $0.5$.  

In the first four settings, we take
\begin{align}
T_i &= \frac{(X_{i2}-1)^2}{4} +\epsilon_i^T; \;\;\epsilon_i^T\iid \mathcal N\l(0, 1\r).
\end{align}

In the fifth setting, we introduce a discontinuity by using
\begin{align}
T_i &=\sign(X_{i1}) \times \frac{(X_{i2}-1)^2}{4} +\epsilon_i^T; \;\;\epsilon_i^T\iid \mathcal N\l(0, 1\r)
\end{align}
where
\begin{align}
    \sign(a)=\begin{cases}
    -1;\;\;a\le 0\\
    1 ;\;\; a>0
    \end{cases}
\end{align}

We then use the following outcome models,
\begin{align}
\textrm{1 Linear: }&\;Y_i=T_i+ X_{i1} + \frac{X_{i2}-1}{4}+ +\epsilon_i\\
\textrm{2 Partially Linear: }&\;Y_i=T_i+ X_{i1} + \frac{(X_{i2}-1)^2}{4}+\epsilon_i\\
\textrm{3 Additive Linear: }&\; Y_i =4\sin(T_i) + X_{i1} + \frac{(X_{i2}-1)^2}{4}+\epsilon_i\\
\textrm{4 Interactive:}&\;Y_i=  4\sin(T_i) \times X_{i1} + \frac{(X_{i2}-1)^2}{4}+\epsilon_i\\
\textrm{5 Discontinuity}&\;Y_i=4  \sin(T_i)\times\sign(X_{i1}) + \frac{(X_{i2}-1)^2}{4}+\epsilon_i
\end{align}
where the the error is independent, identical Gaussian such that the true $R^2$ in the outcome model is $0.5$. 
We consider $n \in \{250, 500, 1000, 2000\}$. The discontinuous model is the one reported in the body of the paper and falls outside the space covered by our model.

We continue to contrast with the kernel regularized least squares model \citep{Hainmueller13} and Generalized Random Forests \citep{athey2019generalized}. We select these two models for comparison because they offer both point estimates and uncertainty estimates for the marginal effect curve, $\tau$.\footnote{We also do not include other candidate approaches. Many existing models focus on uncertainty and estimates for the fitted values, not the marginal effect curve; for example, POLYMARS \citep{Stone1997}, Sparse Additive Models \citep{Ravikumar09}, Bayesian additive regression trees  \citep{chipman2010bart} and boosting \citep{Ridgeway99}, and the SuperLearner  \citep{polleysuperlearner}.  Any of these could have been used for partialing out the covariates; we implemented random forests for simplicity.  Other possible sparse estimators could have included the horseshoe, and Bayesian Bridge \citep{cps2010horseshoe, polson2014bayesian}; we found our EM implementation to offer a more stable estimate than the variational implementation of these.   \citet{catt:farr:feng:19} offer an alternative estimtation strategy, though it does not accommodate more than a handful of covariates.}

\subsubsection{Evaluation Metrics}
We assess methods across two dimensions, each commensurate with our two estimation contributions: point estimation and coverage rates on $\tau$.  For the former, we use the mean absolute error,
\begin{align}
    MAE=\frac{1}{n}\sum_{i=1}^n |\widehat \tau(T_i, X_i) - \tau(T_i, X_i)|
\end{align}
and the sample average coverage probability,
\begin{align*}
SACP=\frac{1}{n}\sum_{i=1}^n
\mathbf 1\left\{\tau(T_i, X_i) \in CB_{\tau, \mathcal D_n}(T_i, X_i)\right\}.
\end{align*}

We note that we expect to get good coverage on the marginal effect, but this will come at the cost of conservative (i.e. wide) confidence intervals.

All simulations were run 200 times.

\subsubsection{Results} 
\begin{figure}[t!]
    \centering
    \includegraphics[width=.8\textwidth]{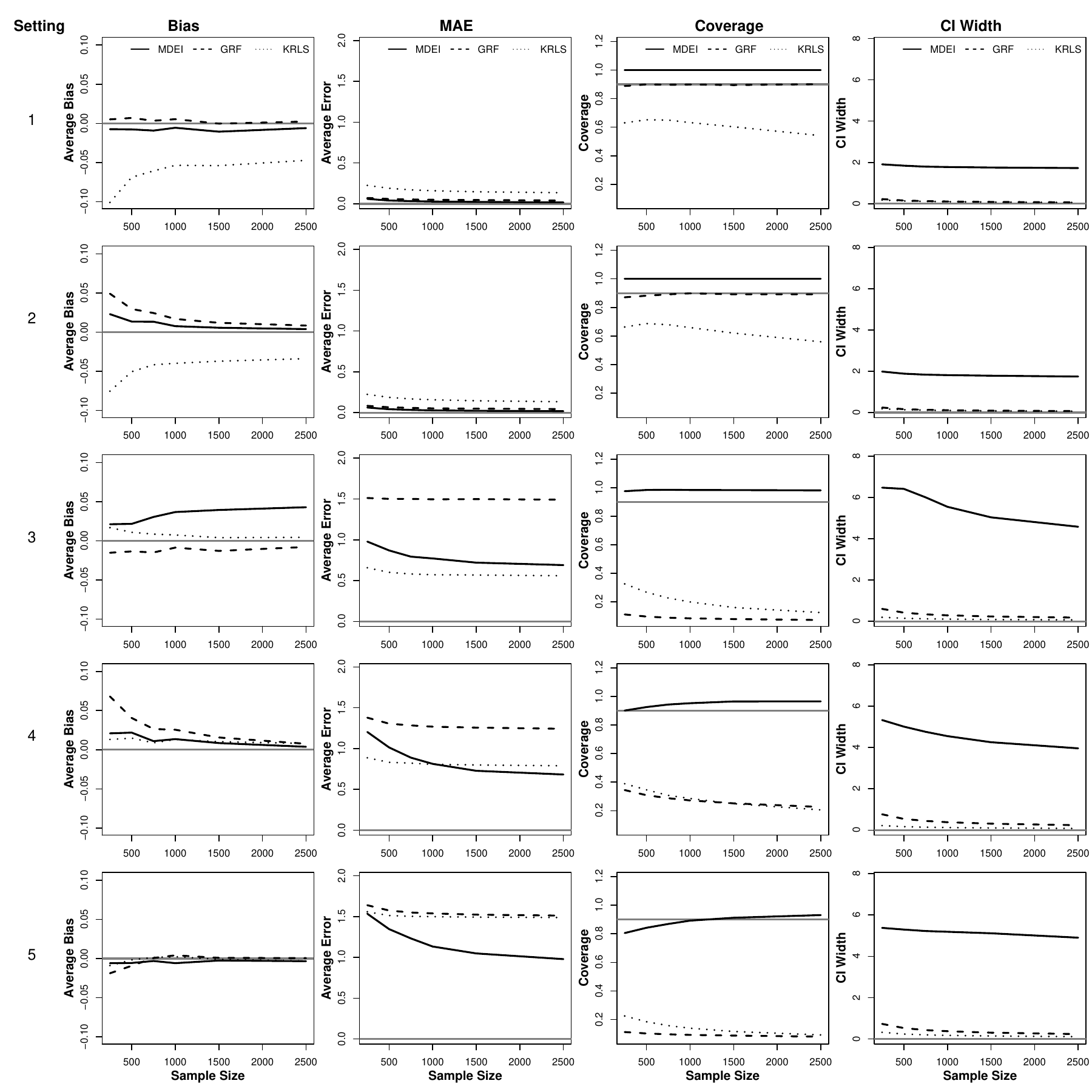}
    \caption{Performance Simulation Results}
    \label{fig:mdesimfig}
\end{figure}

Results from the simulations can be found in Figure \ref{fig:mdesimfig}.  Each row corresponds with our simulation setting, from the additive linear model in row one to the complex, discontinuous model in row 5. In each figure, the $x$-axis shows outcomes by sample size.   The first column presents the bias on the average treatment effect, by method.  Across settings, all methods do well, with GRF doing the best overall.  KRLS misses the average effect in the simplest model, due to its shrinkage, but with any complexity, all models do well.  The second column presents mean absolute error, a measure of accuracy of our estimates over the whole of the curve.  All methods perform well in the simplest settings, but KRLS and MDEI perform the best in the most complex settings.  We suspect that there are conditional mean specifications where any of the methods presented will outperform others; our take way here is that all three methods perform passably well.

The third column, presenting the sample coverage, is the most important.  The horizontal line at $0.9$ is the nominal rate, so values above this line are denote conservative bands and values below it denote an invalid band.  In the simplest settings, all methods are valid, and MDEI is quite wide. This is to be expected, of course, since we construct our bands to be valid even if the model is wrong. As the models get more complex, in rows 3-5, we see that coverage plummets for KRLS and GRF.  Basically, in settings 3 and 4, and especially 5, the confidence bands returned by these method provide little information on the location of the true curve.  The cost of this coverage shows in the last columns, which contains the average width of the interval, by method.  We see that the our confidence intervals are notably wider, as expected.  Narrower bands can be achieved, but at the cost of only covering simple models.

\section{Binary and Categorical Treatment Regimes.}\label{app:bincat}

In the paper we focus on the continuous treatment case. A mature literature examines the case with binary and categorical treatments.  This manuscript does not treat the binary or categorical treatment regime as a separate setting. Instead, we note that our approach carries through to the binary treatment setting.  Rather than modeling the propensity score, or conditional probability of treatment, we instead model the conditional mean of the treatment.  The key distinction is that the former is constrained to fall in $[0,1]$, and the probabilites are used to match or generate inverse probability weights. 

Rather than adjust through matching or weighting, we are instead adjusting the conditional mean.  So, instead of fitting a logistic or probit regression, we are instead fitting using nonlinear least squares.  The benefit is that we are not working with inverse probability weights, which can be unstable, nor relying on distributional assumptions of the treatment variable or outcome. We lose, though, efficiency gains that come from making distributional assumptions.

\end{document}